%% file: main.tex
\documentclass{article}

\usepackage{microtype}
\usepackage{graphicx}
\usepackage{subcaption}
\usepackage{booktabs}
\usepackage{hyperref}
\usepackage{xcolor}
\usepackage{multirow}
\usepackage{enumitem}
\usepackage{makecell}

\usepackage[accepted]{icml2026}

\usepackage{amsmath}
\usepackage{amssymb}
\usepackage{mathtools}
\usepackage{amsthm}
\usepackage{bm}
\usepackage{xcolor}  

\usepackage[capitalize,noabbrev]{cleveref}
\newcommand{\Aref}[1]{Appendix~\ref{#1}}

\usepackage{appendix-toc}

\theoremstyle{plain}

\theoremstyle{definition}

\theoremstyle{remark}

\icmltitlerunning{On the Relationship Between Activation Outliers and Feature Death in Sparse Autoencoders}

\begin{document}

\twocolumn
[
  \icmltitle{On the Relationship Between Activation Outliers and Feature Death in Sparse Autoencoders}

  \icmlsetsymbol{equal}{*}

\begin{icmlauthorlist}
    \icmlauthor{Elana Simon}{inst1}
    \icmlauthor{Etowah Adams}{inst2}
    \icmlauthor{James Zou}{inst1}
\end{icmlauthorlist}

\icmlaffiliation{inst1}{Stanford University}
\icmlaffiliation{inst2}{Columbia University}

\icmlcorrespondingauthor{James Zou}{jamesz@stanford.edu}


  \icmlkeywords{Sparse Autoencoders, Mechanistic Interpretability, Feature Learning, Neural Network Analysis}

  \vskip 0.3in
]

\printAffiliationsAndNotice{}

\begin{abstract}
Sparse autoencoders (SAEs) decompose neural network activations into interpretable features, but
many learned features never activate, a problem called feature death that wastes dictionary
capacity and can reintroduce superposition. Death rates vary dramatically between models: near-zero on GPT-2,
over 70\% on AlphaFold3 with identical configurations. We find that dimension-level activation
outliers (dimensions whose mean magnitude is large relative to per-token variation) cause this
by shifting pre-activations at initialization based on each feature's alignment with the
activation mean. Features anti-aligned with the mean receive permanently negative
pre-activations and never fire. We formalize outlier severity as $\gamma =
\|\boldsymbol{\mu}\|/\|\boldsymbol{\sigma}\|$; it predicts initial death rates (Spearman $\rho
= 0.89$ for dead-by-TopK, $0.82$ for dead-by-ReLU) across 454 model-layer combinations spanning language, vision, protein, and genomic models. Dead
features can revive during training, but recovery requires the SAE bias to learn the
activation mean, a process that is prohibitively slow at high $\gamma$.
Mean-centering (subtracting the activation mean) sidesteps this and eliminates
outlier-induced death across all tested models, confirming the mechanism and providing a
principled basis for when and why this preprocessing step is necessary.
\end{abstract}

\begin{figure*}[t]
\centering
\includegraphics[width=\textwidth]{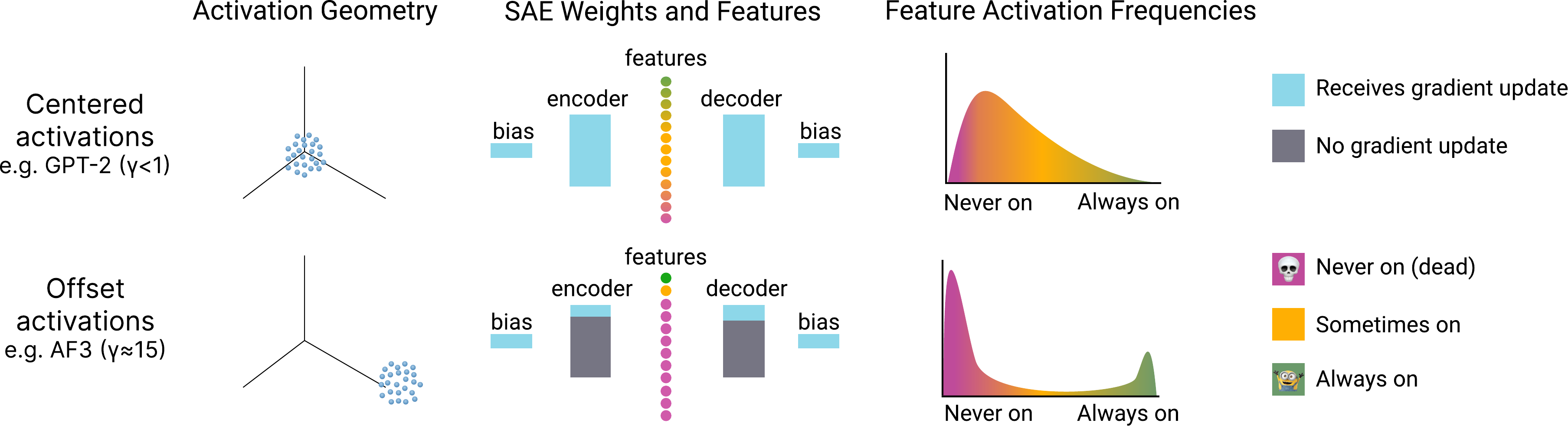}
\caption{
\textbf{Activation outliers predetermine feature fate at initialization.}
(Left) Centered activations (e.g.\ GPT-2, $\gamma < 1$) versus activations
with dimension-level outliers (e.g.\ AF3, $\gamma \approx 15$), where
$\gamma = \|\boldsymbol{\mu}\|/\|\boldsymbol{\sigma}\|$ measures how much
the activation mean dominates per-token variation.
(Middle) In the offset case, only features positively aligned with the
activation mean fire and receive gradient; anti-aligned features are dead
from initialization.
(Right) Centered features have input-dependent activation frequencies;
features from offset activations are primarily always-on or never-on.
}
\label{fig:main}
\end{figure*}

\section{Introduction}

Identical sparse autoencoders (same architecture, same hyperparameters, same auxiliary
losses) produce near-zero dead features on GPT-2 \cite{radford2019} and 72\% dead features on
AlphaFold3 \cite{abramson2024}. Why?

Neural networks are thought to represent more concepts than they have dimensions, encoding them
as overlapping directions in activation space, a phenomenon called superposition
\cite{elhage2022}. Sparse autoencoders (SAEs) attempt to reverse this: they map activations to a
higher-dimensional space and enforce sparsity, learning a dictionary of directions
\cite{bricken2023, huben2024sparse}. Each direction, or \textit{feature}, ideally corresponds to a single interpretable concept;
features that never activate on any input are called \textit{dead features}. With 70\% of
features dead, a
32k-element dictionary represents at most 9.6k concepts, and the surviving features must
compensate, potentially reintroducing the superposition the SAE was meant to resolve. Worse,
death rates vary unpredictably across models and layers (20\% to 80\% within ESM3 alone), making
the effective dictionary size difficult to control.
 
Several techniques attempt to revive dead features by providing gradient signal to dictionary
elements that would otherwise receive none
\cite{jermyn2024ghost, gao2024, bricken2023}. These help on some models but, as we show, remain
ineffective where death is most severe.
 
The pattern does not reduce to model family or domain: not all protein models have high death,
and even within ESM3 \cite{hayes2025simulating}, death rates range from 20\% to 80\% across
layers with the same SAE configuration. Whatever causes death lives in the activations themselves, varying not
just across models but layer by layer within each model.

The cause is what we call \textit{dimension-level activation outliers}. In layers with
high death, certain activation dimensions have mean values that are large relative to their
per-token variation, creating a near-constant offset across all inputs. These differ from the
token-level outliers studied in quantization contexts \cite{sun2024massive, dettmers2022llmint8}:
token-level outliers are properties of individual inputs, while the outliers we study are
properties of dimensions, persistent across all inputs (per-model visualizations in
\Cref{app:fig_skyline_acts_all_models}). These outliers cause dead features
through a geometric mechanism visible at initialization, before any training occurs. Each
feature's pre-activation (the value computed before the sparsity nonlinearity) decomposes into a
constant term, the feature's alignment with the activation mean $\bm{\mu}$, and a varying term
that responds to input content. When outlier dimensions inflate $\|\bm{\mu}\|$, the constant
term dominates: features anti-aligned with the mean can never fire, features strongly aligned
fire on everything, and only features roughly orthogonal to the mean remain input-dependent.
The result is that most features have their fate set at initialization, not learned from data
(Figure~\ref{fig:main}).

We formalize outlier severity as $\gamma = \|\boldsymbol{\mu}\|/\|\boldsymbol{\sigma}\|$, the
ratio of the activation mean's magnitude to the per-token standard deviation magnitude.
Features can die at initialization in two ways: by having permanently negative pre-activations (dead-by-ReLU) or by losing the top-$k$ competition (dead-by-TopK).
From first principles, we derive that the initial dead-feature rate is a closed-form function of $\gamma$ alone, for both pathways. Synthetic experiments confirm this: increasing $\gamma$
monotonically increases death. On real activations from 454 model-layer combinations spanning
language, vision, protein, and genomic models, $\gamma$ predicts initial death rates for both: Spearman $\rho = 0.89$ for dead-by-TopK, and
$\rho = 0.82$ for dead-by-ReLU.

We characterize how dead features can revive during training. Revival has two pathways.
Features that fail to rank in the top-$k$ on any input start ranking once competing features
shrink during training. Features with permanently negative pre-activations recover only as
the bias slowly absorbs the mean offset. This bias-learning pathway is the bottleneck:
at high $\gamma$ the bias has further to travel, and dead features plateau at 75--90\% even
after 2M steps. AuxK \cite{gao2024} accelerates the first pathway but not the second,
which explains why it helps at moderate $\gamma$ but not high $\gamma$.

Mean-centering validates this account directly. If the offset $\bm{\mu}$ causes death,
removing it should eliminate outlier-induced death at initialization, and it does:
mean-centering alone reduces dead features at initialization from 83\% to near zero on ESM3 and from 98\% to
under 5\% on AlphaFold3, without auxiliary losses in either condition. The surviving features
are also higher quality. On ESM3 and DINOv3, two of the highest-death models in our suite, a
mean-centered SAE produces more monosemantic features, and on ESM3 recovers more biological
concepts than a baseline with four times the dictionary size.
Mean-centering has been used in prior SAE work
\cite{bricken2023, gao2024} but
inconsistently and without a clear rationale for when it matters. Our analysis provides one:
$\gamma$ identifies models where centering is necessary, and the recovery analysis explains why
training alone cannot compensate in time.

Mean-centering does not eliminate all feature death. A few layers, primarily in protein and
genomic models, retain residual death from a separate geometric cause: when activation
variance is concentrated in a small number of directions, most features cannot win the
top-$k$ competition regardless of the mean. PCA whitening equalizes variance across
directions and eliminates this residual death (\Aref{sec:app_spectral}).

\textbf{Contributions.}
\begin{enumerate}[topsep=0pt, itemsep=1pt, parsep=0pt]
    \item \textbf{Outlier severity diagnostic.} We derive $\gamma =
    \|\boldsymbol{\mu}\|/\|\boldsymbol{\sigma}\|$ from first principles. Analytically, $\gamma$
    determines the fraction of features with permanently negative pre-activations at
    initialization. The diagnostic matches synthetic experiments and correlates with death
    rates across 454 real model-layer combinations for both pathways (Spearman $\rho = 0.89$
    and $\rho = 0.82$) with no fitting, letting practitioners assess
    SAE trainability before committing compute.

    \item \textbf{Geometric cause of feature death.} We find that dead features in high-$\gamma$
    models are a geometry problem, not a training dynamics problem. Outlier dimensions inflate
    the activation mean, creating input-independent pre-activation shifts that predetermine
    feature fate at initialization. This explains why the same SAE configurations produce
    near-zero dead features on GPT-2 but over 70\% on AlphaFold3: the difference is not in the
    SAE but in the activation geometry.

    \item \textbf{Recovery dynamics.} We characterize how dead features revive during training,
    finding two pathways with different timescales. Features that fail to rank in the top-$k$
    on any input start ranking as competing features shrink. Features with permanently negative
    pre-activations recover only as the bias slowly absorbs the mean offset, a process
    that requires prohibitively long training at high $\gamma$. AuxK helps the first pathway
    but not the second.

    \item \textbf{Validation via mean-centering.} Mean-centering eliminates outlier-induced
    death across all tested models, confirming the geometric account. It also provides a
    principled basis for when this preprocessing step is necessary (high $\gamma$) and why
    existing revival techniques cannot substitute for it.
\end{enumerate}

\section{Background}

\paragraph{Sparse autoencoders.}

Neural networks appear to encode more concepts than they have dimensions, superimposing them as overlapping directions in activation space \citep{elhage2022}. Sparse autoencoders (SAEs) try to undo this by treating activations as mixtures of unknown basis elements and solving for the basis \citep{olshausen_emergence_1996}, using a learned encoder that maps activations to sparse codes in a single feedforward pass \citep{bricken2023, huben2024sparse}. Concretely, an SAE maps activations $\mathbf{x} \in \mathbb{R}^d$ to a higher-dimensional latent space $\mathbf{z} \in \mathbb{R}^n$ (with $n > d$), enforces sparsity on $\mathbf{z}$, and reconstructs $\mathbf{x}$:
{%
\setlength{\abovedisplayskip}{2pt}
\setlength{\belowdisplayskip}{2pt}
\setlength{\jot}{2pt}
\begin{align*}
\mathbf{z}_{\text{pre}} &= \mathbf{W}_{\text{enc}} (\mathbf{x} - \mathbf{b}) + \mathbf{b}_{\text{enc}} \\
\mathbf{z} &= \sigma(\mathbf{z}_{\text{pre}}) \\
\hat{\mathbf{x}} &= \mathbf{W}_{\text{dec}}^\top \mathbf{z} + \mathbf{b}
\end{align*}
}%
Here $\sigma$ is a sparsity-inducing nonlinearity and $\mathbf{b} \in \mathbb{R}^d$ is the \textit{bias}, which centers the input before encoding and recenters the output after decoding. The decoder columns $\mathbf{W}_{\text{dec}}$ form the learned dictionary: each column is a direction in activation space, and the sparse code $\mathbf{z}$ indicates which directions are active for a given input.

The architectures differ mainly in how they enforce sparsity. ReLU SAEs use $\sigma = \text{ReLU}$ with an L1 penalty on $\mathbf{z}$. TopK SAEs apply ReLU and then keep only the $k$ largest positive pre-activations, zeroing the rest, so exactly $k$ features fire per input \citep{gao2024}. JumpReLU SAEs learn per-feature thresholds, allowing adaptive sparsity \citep{rajamanoharan2024jumpingaheadimprovingreconstruction}. We focus on TopK SAEs, but the core findings hold across architectures (\Aref{sec:d_extended_res}).

\paragraph{Dead features and revival methods.}

A dead feature is a dictionary element that never activates on any input. Because dead features receive zero gradient on their encoder weights, they cannot update toward useful directions: they are stuck. This can happen in two ways. First, any architecture that applies ReLU (or an equivalent threshold) will zero out a feature whose pre-activation is always negative, regardless of sparsity selection. Second, TopK SAEs introduce a competition-based pathway: a feature with positive pre-activations can still die if it never ranks among the top-$k$ on any input. Other architectures have analogous competition effects (JumpReLU features can be pushed below their learned thresholds), but TopK's hard selection makes this pathway especially stark.

Several methods try to revive dead features by injecting gradient signal where there would otherwise be none. AuxK \citep{gao2024} adds an auxiliary loss based on how well dead features reconstruct the residual error. Ghost gradients \citep{bricken2023} backpropagate through a reconstruction computed from dead features. Resampling \citep{bricken2023} periodically reinitializes dead feature weights toward directions with high reconstruction error. These methods treat dead features as a training dynamics problem: the features are stuck, so push them. Comparatively little attention has gone to why some models produce massive death rates in the first place; one recent exception, \citet{wang2025attentionlayersaddlowdimensional}, links death to low-rank structure in attention activations. Our focus is a different geometric cause: dimension-level activation outliers.

\section{Dead Feature Rates Vary Dramatically Across Models}

\begin{figure}[h]
\centering
\includegraphics[width=\columnwidth]{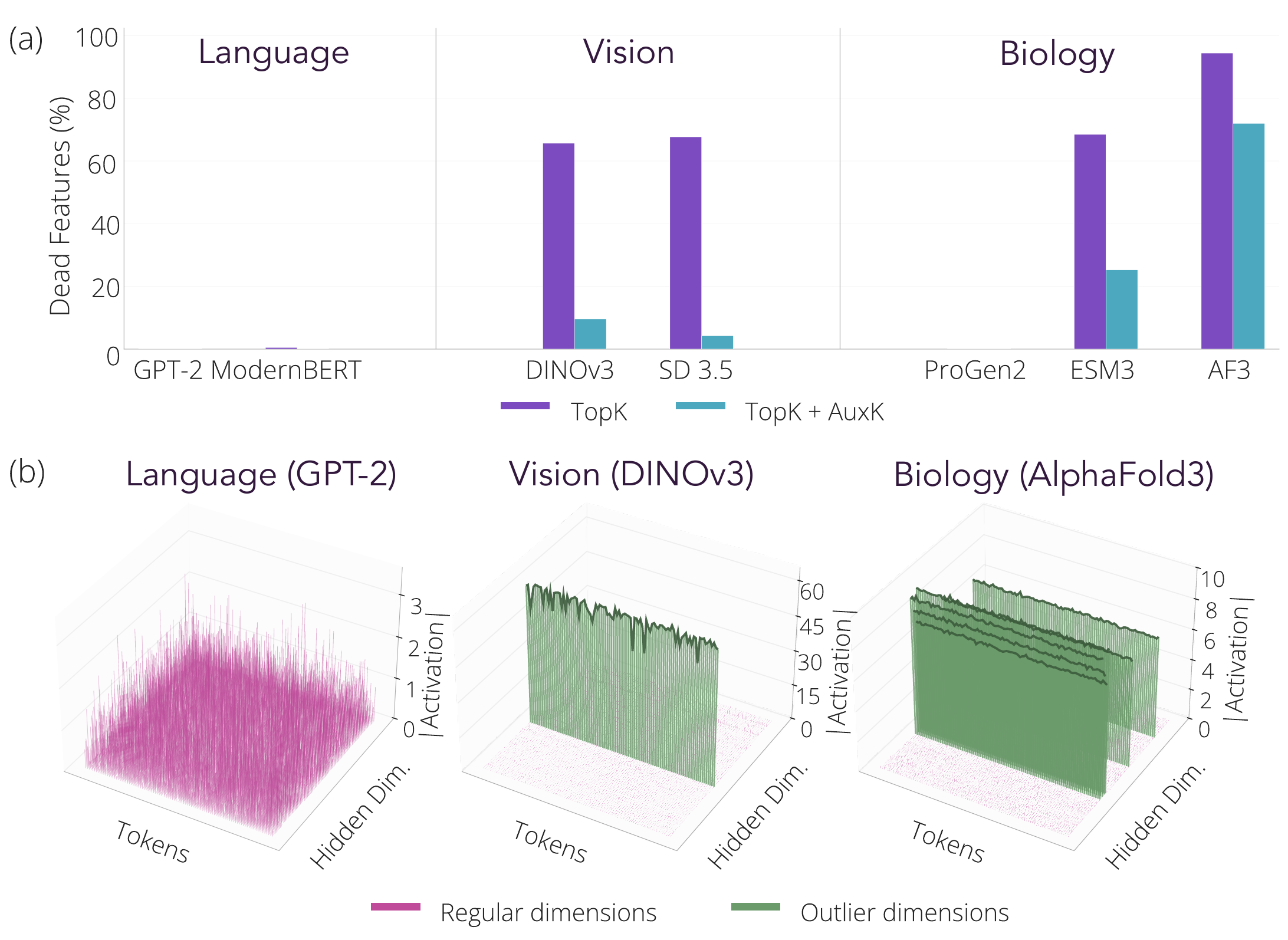}
\caption{
\textbf{Dead feature rates vary dramatically across models and coincide
with dimension-level activation outliers.}
(a) Dead feature rates for TopK SAEs across a subset of language, vision,
protein, and genomic models (additional models in Figure~\ref{fig:extended_all_models}). AuxK helps but does
not resolve death where it is most severe.
(b) Activation magnitude landscapes (LayerNorm-normalized per token) for
three models. Green highlights dimensions with high mean and low per-token
variance. High-death models are distinguished by dimensions whose magnitude
is \emph{consistently large} across tokens, not merely high-variance.
}
\label{fig:death_patterns}
\end{figure}

When we train SAEs on activations from a range of models and modalities, dead feature rates vary dramatically, from under 5\% to over 95\% (\Cref{fig:death_patterns}a). SAEs were originally established as a tool for interpreting language models \citep{huben2024sparse, bricken2023} but have since been applied across many domains including vision \citep{gorton2024missingcurvedetectorsinceptionv1,joseph2025steeringclipsvisiontransformer} and biology \cite{simon2025interplm, adams2025mechanistic, brixi2026evo2}. We train TopK SAEs with identical architecture, dictionary size, sparsity, and learning-rate sweep on source models spanning language, vision, protein, and genomic, using one SAE per model trained on a representative (mid-network) layer (full model list and hyperparameters in \Aref{app:exp_setup}). For all language models in our suite, we could train SAEs with no dead features; for many protein and vision models, even the best configuration left the majority dead. AuxK reduces death on some models but cannot bring the worst cases below 75\%.

Among the models we examine, biology and vision models tend toward higher death than language models. But the variation within a single model can be just as large: ESM3 ranges from over 80\% dead in early layers to under 20\% in later layers, with the same SAE and same hyperparameters. This implies the cause is something about each layer's activation distribution that some model families produce more often than others.

High-death layers share a specific activation pattern: certain dimensions take on large, near-constant activations across every token. This high-mean-low-variance signature differs from the per-token spikes typically studied as activation outliers (often measured by kurtosis; e.g., \citealp{outlierfeatures2024neurips}). \citet{lu2024} previously identified the same pattern in ESMFold. Compare GPT-2, where no dimension dominates, to ESM3 and AlphaFold3, where a handful of dimensions tower above the rest on every input (\Cref{fig:death_patterns}b; per-model visualizations in \Cref{app:fig_skyline_acts_all_models}).

\section{Why Outliers Cause Death at Initialization}
\label{sec:init_death}

One might expect outliers to cause dead features indirectly: distorting the loss, producing bad gradients, eventually breaking the encoder. But we observe something more direct: most features are already dead at initialization, before any gradients flow. The problem is not corrupted training dynamics, but rather a corrupted starting point.

\begin{figure}[h]
\centering
\includegraphics[width=\columnwidth]{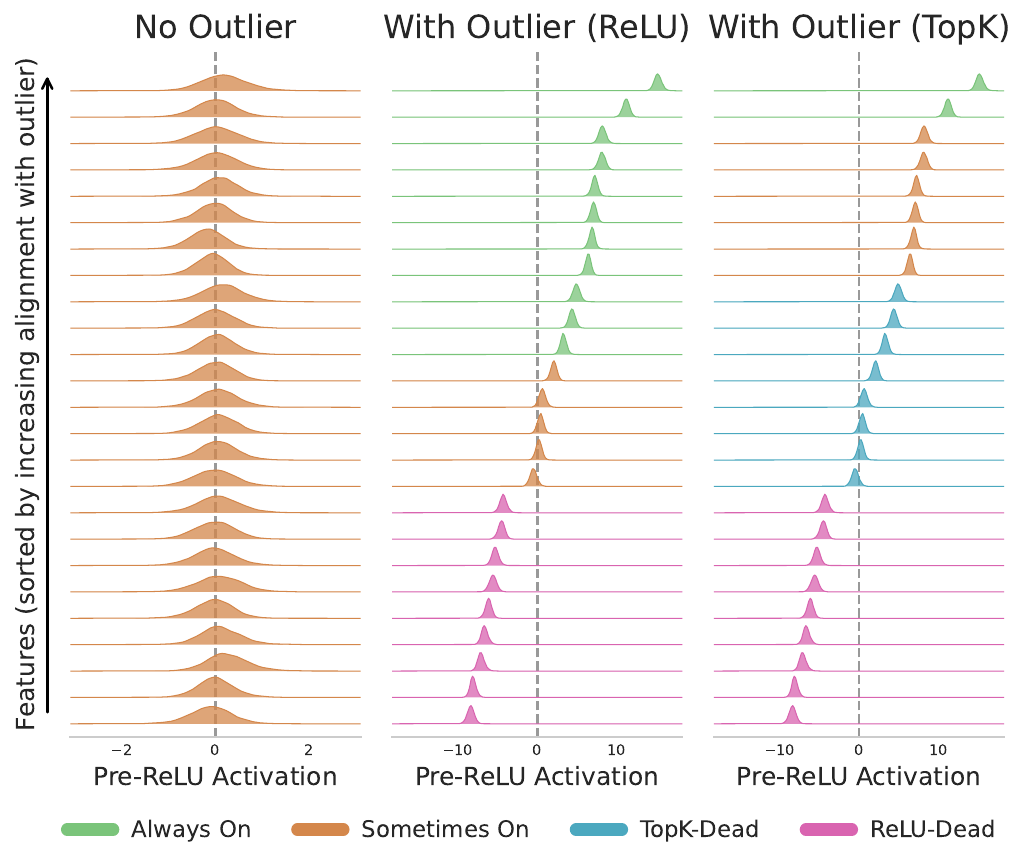}
\caption{
\textbf{A feature's alignment with the activation mean determines its fate
at initialization.}
Each row is one feature sorted by alignment with $\boldsymbol{\mu}$;
ridgelines show its pre-activation distribution across inputs.
Left (no outliers): all features are input-dependent.
Middle (with outliers, ReLU SAE): anti-aligned features have permanently
negative pre-activations (dead-by-ReLU); strongly aligned features fire on
everything; only roughly orthogonal features remain input-dependent.
Right (with outliers, TopK SAE): additional features with moderate positive
pre-activations die because they never rank in the top $k$
(dead-by-TopK).
}
\label{fig:pathways}
\end{figure}

\subsection{Features anti-aligned with an outlier-inflated mean are initialized dead}

Consider activations where one dimension has value ${\sim}1000$ regardless of input, while other dimensions vary around zero with standard deviation ${\sim}1$. At initialization the biases are zero, so the pre-activation for feature $i$ is $z_i = \mathbf{w}_i \cdot \mathbf{x}$. If $w_{i,0} = +0.01$, dimension 0 alone contributes $+10$, swamping the varying dimensions. This feature fires on every input. If $w_{i,0} = -0.01$, the contribution is $-10$, and the feature never fires. Only features with $w_{i,0} \approx 0$ remain sensitive to input content.

This decomposition applies to arbitrary activation distributions, not just the single-outlier toy case. Decomposing the activation for an individual token as $\mathbf{x} = \bm{\mu} + (\mathbf{x} - \bm{\mu})$:
{%
\setlength{\abovedisplayskip}{1pt}%
\setlength{\belowdisplayskip}{3pt}%
\setlength{\abovedisplayshortskip}{1pt}%
\setlength{\belowdisplayshortskip}{2pt}%
\begin{equation*}
z_i = \underbrace{\mathbf{w}_i \cdot \bm{\mu}}_{\text{shift (constant)}} + \underbrace{\mathbf{w}_i \cdot (\mathbf{x} - \bm{\mu})}_{\text{signal (varies with input)}}
\end{equation*}
}

Whether a feature responds to input content or has its fate predetermined at initialization depends on which term dominates. The ratio $\gamma = \|\bm{\mu}\|/\|\bm{\sigma}\|$ quantifies this: high $\gamma$ means shifts dominate signals.

This produces two distinct failure modes (\Cref{fig:pathways}). \textbf{Dead-by-ReLU:} features with large negative shifts ($\mathbf{w}_i \cdot \bm{\mu} \ll 0$) have pre-activations that no input can push above zero. \textbf{Dead-by-TopK:} features with moderate positive shifts pass ReLU but lose the TopK competition to features with large positive shifts; the same features win every selection. Both pathways have predictable extremes. Under symmetric initialization, roughly half of features align positively with outliers and half negatively, so dead-by-ReLU approaches 50\% at high $\gamma$. For TopK, only the $k$ features most positively aligned with $\boldsymbol{\mu}$ ever win the competition, so dead-by-TopK approaches $1 - k/n$ (e.g. 99.2\% at $k = 64$, $n=8192$). Beyond these extremes, we can derive analytically how $\gamma$ determines the death rate at any outlier severity.

\subsection{Outlier Severity Predicts Death Rates Analytically}
\label{sec:analytical}

A feature dies when its negative shift exceeds the largest signal fluctuation across $N$ evaluation samples; as shift and signal are projections of random unit vectors onto fixed directions, we can use high-dimensional probability rules to analytically compute the probability of death-by-ReLU based on outlier severity (step by step derivation in \Aref{sec:derivation}):
{%
\setlength{\abovedisplayskip}{1pt}%
\setlength{\belowdisplayskip}{3pt}%
\setlength{\abovedisplayshortskip}{1pt}%
\setlength{\belowdisplayshortskip}{2pt}%
\begin{equation*}
P(\text{dead-by-ReLU}) = \Phi\left(\frac{-C}{\gamma}\right)
\end{equation*}
}
where $\Phi$ is the standard normal CDF and $C = \Phi^{-1}(1 - 1/N)$ depends only on the number of evaluation samples ($C \approx 4.26$ for $N = 100{,}000$).

For dead-by-TopK the same setup applies, but the survival bar is higher: a
feature has to land in the top $k$ pre-activations out of $n$ on at least
one input. When $\gamma$ is large the bar is approximately fixed across
inputs: the spread of shifts across features is roughly $\gamma$ times the spread of signals,
so the top $k$ are essentially the $k$ features with the largest
shifts no matter what the input is. The bar a feature must clear is then
the $(1 - k/n)$ quantile of the shift distribution: the value such that
only $k$ out of $n$ features have larger shifts. Substituting this for
zero in the dead-by-ReLU derivation gives:
{%
\setlength{\abovedisplayskip}{1pt}%
\setlength{\belowdisplayskip}{3pt}%
\begin{equation*}
P(\text{dead-by-TopK}) \;\approx\; \Phi\!\left(t_k - \frac{C}{\gamma}\right),
\quad t_k = \Phi^{-1}\!\left(1 - \frac{k}{n}\right)
\end{equation*}
}%
The dead-by-ReLU formula is the $t_k = 0$ special case (full derivation in \Aref{sec:topk_derivation}).

\begin{figure}[h]
\centering
\includegraphics[width=0.9\columnwidth]{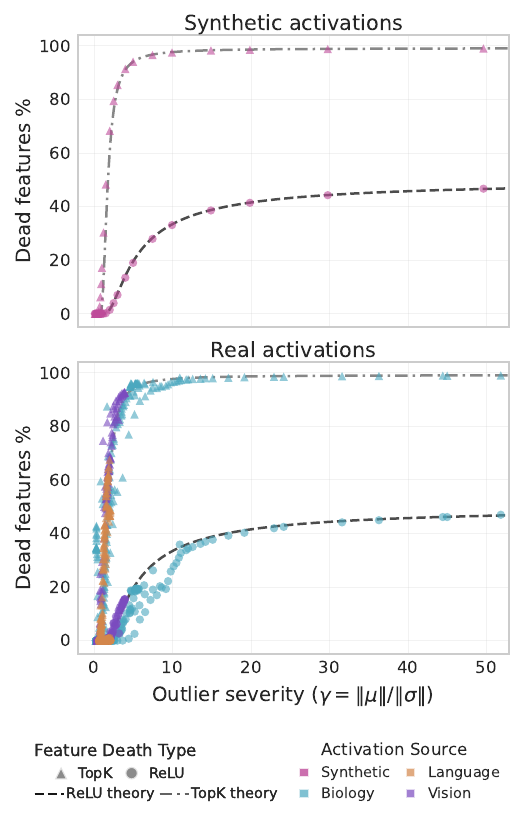}
\caption{
\textbf{$\gamma$ analytically predicts dead feature rates for both death
pathways.}
(a)~Synthetic activations with controlled $\gamma$: dead features increase
monotonically (Spearman $\rho = 1.0$). The ReLU theory curve
$\Phi(-C/\gamma)$ matches dead-by-ReLU rates (circles); the TopK theory
curve $\Phi(t_k - C/\gamma)$ tracks dead-by-TopK (triangles), becoming tight
for $\gamma > 3$. The dotted line at 99.2\% marks the $1-k/n$ asymptote.
(b)~454 real model-layer combinations spanning language, vision, protein, and genomic.
The same relationships hold: Spearman $\rho = 0.82$ (dead-by-ReLU) and
$0.89$ (dead-by-TopK).
}
\label{fig:gamma_combined}
\end{figure}

\subsection{Does $\gamma$ Predict Death in Practice?}

\textbf{Synthetic experiments establish causality.} We generate activations with controlled $\gamma$ (single dominant outlier dimension; details in \Aref{app:synthetic_data_details}) and train SAEs. Dead features increase monotonically with $\gamma$, with perfect rank correlation (Spearman $\rho = 1.0$). The theoretical curve $\Phi(-C/\gamma)$ matches observed dead-by-ReLU rates closely (\Cref{fig:gamma_combined}). Dead-by-ReLU plateaus near 50\% at high $\gamma$; dead-by-TopK continues rising toward $1 - k/n$ as only the most positively-aligned features survive.

\textbf{Real activations show the same pattern.} Across 454 model-layer
  combinations spanning language, vision, protein, and genomic models, $\gamma$ predicts
   death rates with Spearman $\rho = 0.82$ for dead-by-ReLU and $\rho = 0.89$
  for dead-by-TopK. These correlations are particularly notable given that the derivation assumes signal projections are Gaussian, which fails when activations are heavy-tailed (\Aref{sec:heavy_tails}), and that other geometric factors can also drive death (\Aref{sec:app_spectral}). The correlation holds both across and within model families
   (\Cref{fig:gamma_combined}). On real data we apply per-token LayerNorm (LN) \citep{ba2016layernorm} before computing $\gamma$. This preprocessing isolates the per-token outlier structure that drives feature death from between-token scale variation (largest in vision transformers), which would otherwise inflate $\|\bm{\sigma}\|$ uniformly across dimensions and cause raw $\gamma$ to understate outlier severity (\Aref{sec:per_token_scale}).

The formula slightly over-predicts dead-by-ReLU on some layers. The derivation treats signal projections as Gaussian, so across $N$ samples the largest signal concentrates near $C \approx 4.26$ standard deviations: a feature whose shift is more negative than this can never be pushed above zero. Real activations sometimes have heavier tails, and the largest sample then reaches well beyond $C$. Features with moderate negative shifts that the Gaussian formula declares dead are then occasionally rescued by these tail excursions. We diagnose this directly from per-dimension kurtosis of the activations themselves, which is cheap to compute and tracks where the formula breaks down (\Aref{sec:heavy_tails}, \Cref{fig:kurtosis_mechanism}).


\begin{figure}[t]
\centering
\includegraphics[width=\columnwidth,keepaspectratio]{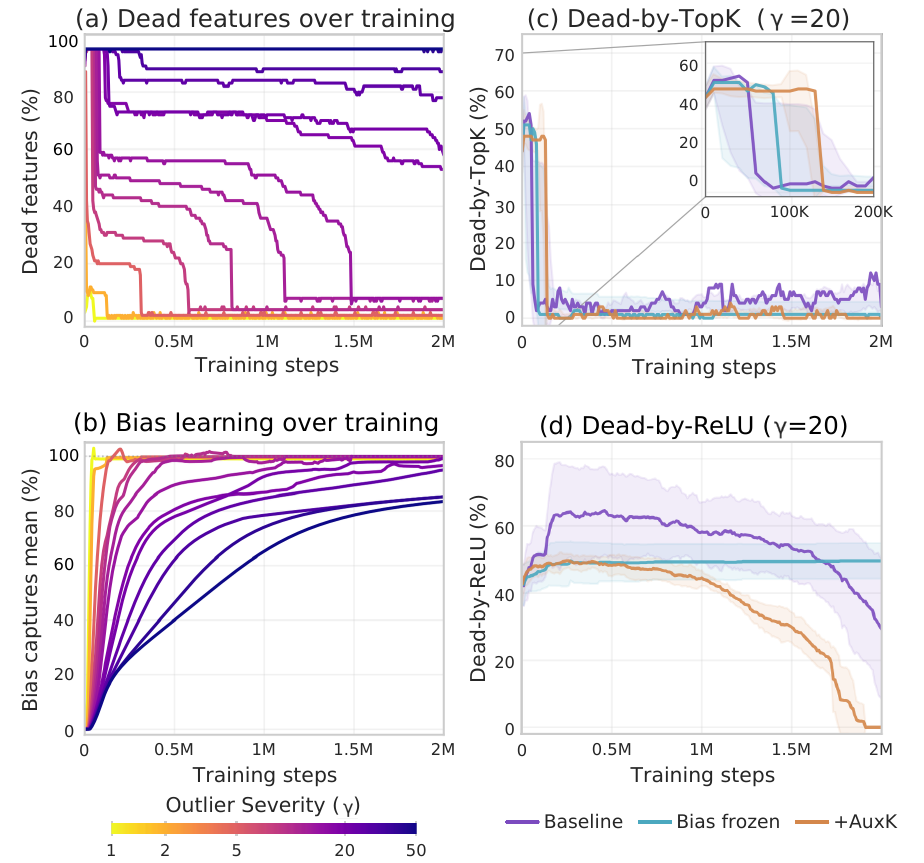}
\caption{
\textbf{Recovery from outlier-induced death is bottlenecked by slow bias
learning.}
All panels use synthetic activations with $\gamma$ set directly; the same
two-phase pattern holds on real models (\Cref{fig:real_dynamics}).
(a) Dead feature recovery slows as $\gamma$ increases.
(b) Bias convergence to $\boldsymbol{\mu}$ slows as $\gamma$ increases.
(c) Dead-by-TopK at $\gamma = 20$ drops steeply around 100K steps (inset)
regardless of condition.
(d) Dead-by-ReLU at $\gamma = 20$ is much slower: frozen bias prevents
recovery entirely, and AuxK prevents the early spike but does not speed
the underlying decline.
Shaded bands: $\pm 1$ s.d.\ over 10 seeds.
}
\label{fig:training}
\end{figure}

\section{Dead Features Revive During Training, but Recovery Is Slow}
\label{sec:how-recovery-works}

If outlier-induced death is just a mean offset, and SAEs have bias terms that can
learn offsets, shouldn't training eventually fix itself? It does, but slowly.
We study these dynamics with synthetic activations, which let us set $\gamma$
directly, and ablate individual components (bias, AuxK, sparsity competition)
in isolation; the same dynamics hold on real models (\Cref{fig:real_dynamics}).
Figure~\ref{fig:training}a tracks dead features over training across $\gamma$
values. At low $\gamma$, recovery finishes within a few hundred thousand steps.
At high $\gamma$ ($\geq 30$), dead features plateau at 75--90\% even after
2M steps. We find that the two death pathways (dead-by-ReLU and dead-by-TopK) recover through different
mechanisms, one fast and one slow, and the slow one is bottlenecked by the
bias learning the activation mean, which takes longer for larger means.

\subsection{The two death pathways are resolved by different parameters}
\label{sec:two_stages}

Decomposing dead features at $\gamma = 20$ by death pathway (\Cref{fig:training}c,d) reveals two very different dynamics. Dead-by-TopK features (\Cref{fig:training}c) revive within ${\sim}$200K steps in all three conditions: the baseline, an ablation where the bias parameter is held frozen during training, and +AuxK. This produces the sharp early drop visible in \Cref{fig:training}a. Dead-by-ReLU features (\Cref{fig:training}d) decline much more slowly: the baseline continues to drop across the remaining 1.8M steps but does not fully reach zero by the end of training, and under frozen bias they remain permanently dead (individual feature trajectories in \Aref{app:per_feature}).

To understand why the two pathways have such different revival timescales, we look at how each one actually revives. Dead features in either pathway receive no gradient on their own encoder weights, since TopK or ReLU zeros their decoder contribution and no signal flows back. Revival therefore requires other parameters to change. For dead-by-ReLU features, the only parameter that can shift a stuck-negative pre-activation above zero is the bias, so these features revive only as the bias updates. For dead-by-TopK features, the dominant mechanism is encoder weights of alive features changing: as those alive features (which do receive gradient) reduce their own activations during training, dead-by-TopK features rise into the top-$k$. The bias could in principle also reshuffle the TopK rankings, but empirically does not drive revival. Freezing the bias during training makes the asymmetry visible: TopK revival proceeds almost unchanged, but dead-by-ReLU features remain permanently dead.

A side effect of the TopK revival mechanism shows up in \Cref{fig:training}d: dead-by-ReLU rates \textit{increase} during the first ${\sim}$200K steps, the same window where TopK revival is happening. When alive features reduce their activations to free up TopK slots, some get pushed below zero in the process and become dead-by-ReLU. This collateral death increases with $\gamma$ (\Aref{app:decomp_gamma}), and explains why the sharp early drop in \Cref{fig:training}a becomes less visible at high $\gamma$: TopK revival is partially offset by new ReLU deaths, so total dead count stays relatively flat despite active turnover underneath. This collateral death phenomenon will also help us understand AuxK's effect on dead-by-ReLU recovery in \Cref{sec:auxk}; see \Aref{app:recovery} for further analysis of these recovery dynamics.

\subsection{Bias learning is the bottleneck}
\label{sec:bias_bottleneck}

Why does dead-by-ReLU recovery scale so much worse with $\gamma$ than dead-by-TopK? It turns out bias learning is the bottleneck, and it slows with $\gamma$. \Cref{fig:training}b shows bias convergence to the mean over training (one curve per $\gamma$): at $\gamma \leq 5$, the bias reaches ${\sim}$99\% of $\boldsymbol{\mu}$ within ${\sim}$200K steps; at $\gamma \approx 20$, only ${\sim}$90\% after 2M steps; at $\gamma \geq 30$, only 50--70\%.

Features, by contrast, capture $\boldsymbol{\mu}$ much faster than the bias. Feature weights are multiplied by inputs, so small weight changes have effects that scale with input magnitude; the bias is added directly, with effects independent of input scale. As $\|\boldsymbol{\mu}\|$ grows, features can keep up via small weight changes while the bias takes proportionally longer. Once alive features capture $\boldsymbol{\mu}$, they reduce the reconstruction residual that drives the bias gradient, slowing bias learning further (\Aref{app:lever_arm}).

\subsection{AuxK prevents collateral death but does not speed up bias learning}
\label{sec:auxk}

Comparing \Cref{fig:training}c and \Cref{fig:training}d, we see AuxK's most
visible benefit is on dead-by-ReLU (\Cref{fig:training}d), not
dead-by-TopK (\Cref{fig:training}c). TopK revival proceeds on a similar
timescale regardless of whether AuxK is used. This is surprising. AuxK's
auxiliary loss applies ReLU before computing reconstructions from dead features,
so features with negative pre-activations receive no gradient from it. Given
that we observe bias learning as the bottleneck
for dead-by-ReLU recovery, a natural guess is that AuxK works by speeding up
bias learning. But that's not what happens.

AuxK does not noticeably speed up bias convergence
(\Aref{app:lever_arm}); rather, it reduces collateral death. Recall from Section~\ref{sec:two_stages} that TopK revival creates new
dead-by-ReLU features: once features aligned with the mean
learn to reconstruct it, they start decreasing their activations; the threshold drops,
opening TopK slots but also pushing some active features below zero. AuxK
provides gradient to dead-by-TopK features, helping them stabilize above zero
rather than crossing into dead-by-ReLU. The reduction in dead-by-ReLU that AuxK
achieves comes from \textbf{preventing collateral death-by-ReLU}, not from recovering features that were dead-by-ReLU from the start.

This provides one explanation for why AuxK's benefit depends on $\gamma$. At moderate $\gamma$
(10--20), collateral death accounts for a large share of persistent
dead-by-ReLU features, so preventing it is the difference between full recovery
and permanent death. At high $\gamma$ ($\geq 30$), most dead-by-ReLU features
were dead from initialization, born anti-aligned with $\boldsymbol{\mu}$, not
created during threshold collapse. Preventing collateral death addresses a
shrinking fraction of the total problem, so AuxK provides diminishing benefit
(\Aref{app:decomp_gamma}).

The common thread across all of these dynamics is $\boldsymbol{\mu}$: it
predetermines feature fate at initialization, creates both death pathways, and
imposes a recovery timescale that grows with its magnitude. Rather than waiting
for the bias to slowly learn $\boldsymbol{\mu}$ during training, we can
initialize it with the mean directly.

\section{Mean-centering eliminates outlier-induced death and improves feature quality}
\label{sec:validation}

\begin{figure}[h]
\centering
\includegraphics[width=\columnwidth]{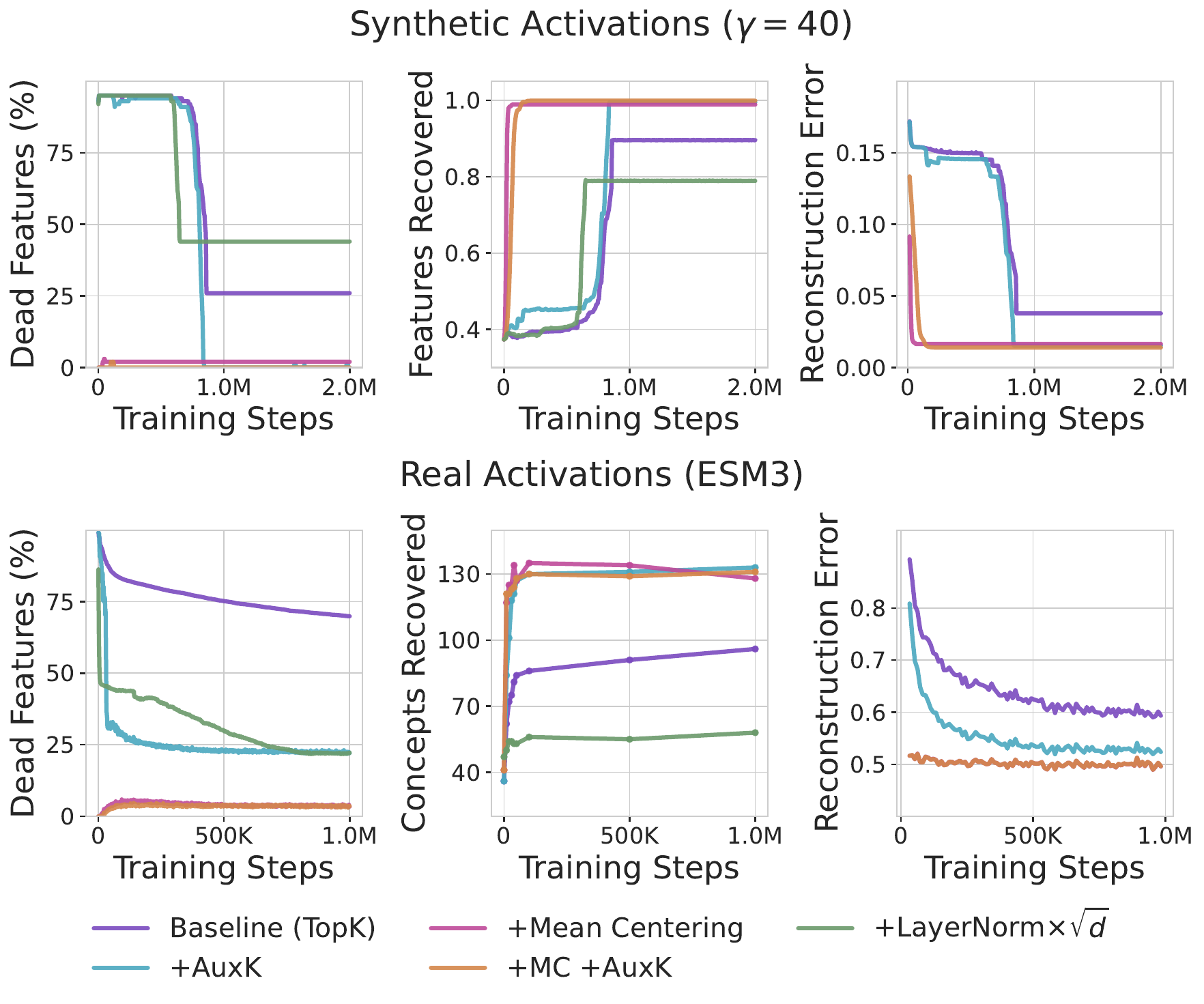}
\caption{
\textbf{Mean-centering sidesteps the recovery bottleneck: features are
never dead to begin with.} Top: synthetic data ($\gamma = 40$); mean-centering achieves near-zero
death, near-perfect recovery of ground-truth features, and lower reconstruction error.
Bottom: ESM3 layer~24 ($\gamma \approx 8$); baseline reaches around $75\%$
dead by end of training, AuxK plateaus ${\sim}25\%$, mean-centering stays near zero and
recovers more biological concepts than baseline.
}
\label{fig:centering_synthetic}
\end{figure}

When we initialize $\textbf{b}$ with $\bm{\mu}$ (effectively mean-centering our activations), the pre-activation becomes:
{
\setlength{\abovedisplayskip}{4pt}
\setlength{\belowdisplayskip}{4pt}
\setlength{\abovedisplayshortskip}{4pt}
\setlength{\belowdisplayshortskip}{2pt}
\begin{equation*}
z_i = \mathbf{w}_i \cdot (\mathbf{x} - \bm{\mu}) + b_{\text{enc}}
\end{equation*}
}
The shift term $\mathbf{w}_i \cdot \bm{\mu}$ vanishes and all features have pre-activations centered around zero, varying only with input content. No features are born dead from this mean offset.

Rather than subtracting $\bm{\mu}$ at runtime, we fold it into bias initialization: set the bias to the activation center (we use the geometric median by default, which works better than the arithmetic mean on certain models; details and per-model comparison in \Aref{app:gm_vs_am}). This is mathematically equivalent to runtime centering but requires no additional runtime computation.

\subsection{Mean-centering eliminates dead features at initialization}

If our explanation of outlier-induced death is correct, mean-centering should eliminate it from the start. We test this on synthetic data with controlled $\gamma = 40$, and on our representative middle layer from ESM3 (L24, $\gamma \approx 8$).

On synthetic data (\Cref{fig:centering_synthetic} top), mean-centering achieves near-zero dead features throughout training and reduces reconstruction error by an order of magnitude; the baseline never recovers within 2M steps (${\sim}8$B tokens), and AuxK reaches similar dead-feature counts only via threshold collapse around 800K steps (the same dynamics from \Cref{sec:how-recovery-works}, accelerated because AuxK provides gradient to otherwise-stuck features). On ESM3 layer 24 (\Cref{fig:centering_synthetic} bottom), baseline dead features reach around 75\% by end of training; AuxK drops to ${\sim}$25\% but plateaus; mean-centering starts and stays near zero. LayerNorm with $\sqrt{d}$ rescaling \citep{templeton2024scaling} reduces ESM3 death to ${\sim}20\%$ but recovers fewer concepts than baseline, and does not match mean-centering on synthetic data.

\Cref{fig:extended_all_models} summarizes results across all source models, each trained on its representative middle layer for 1M steps: mean-centering eliminates outlier-induced death consistently, with improvements proportional to $\gamma$. Low-$\gamma$ models (GPT-2, Pythia) show minimal change; high-$\gamma$ models (ESM3, AlphaFold3) show dramatic reductions.

Mean-centering also reduces sensitivity to learning-rate choice: baseline dead-feature rates vary widely across the LR sweep, while mean-centered rates stay consistently low. While we focus on TopK SAEs (where we can easily fix sparsities across models), ReLU and JumpReLU SAEs also benefit from mean-centering (\Aref{sec:d_extended_res}).

\subsection{Surviving features are more useful}
\label{sec:more_useful}

Beyond reducing dead-feature counts, mean-centering produces features that better recover the structure we expect SAEs to find. Two settings let us compare SAE features to \emph{expected} ones: synthetic data with ground-truth feature directions we constructed, and protein models where features should encode known biological structure. In both, mean-centered SAEs recover more of the expected features.

On synthetic data, mean-centered SAE features align with the ground-truth feature directions much more closely than baseline. The mean maximum cosine similarity (MMCS) between learned and ground-truth features is 0.97 for mean-centered SAEs vs.\ 0.38 for baseline (full metric definition in \Aref{app:metrics}).

On our representative middle layer of ESM3, mean-centered SAEs capture substantially more known biological concepts (\Cref{tab:concepts_by_dict}; methods in \Aref{app:metrics}). Mean-centering outperforms a 4$\times$ larger dictionary here: a mean-centered SAE with 2048 features captures more concepts (100) than a baseline SAE at 8192 (73), at a fraction of the compute.

\begin{table}[h]
\centering
\setlength{\tabcolsep}{4pt}
\footnotesize
\begin{tabular}{lcc}
\toprule
Dict size & Baseline & \makecell{+Mean\\Center} \\
\midrule
2048 & 61 & 100 \\
4096 & 69 & 121 \\
8192 & 73 & 127 \\
\bottomrule
\end{tabular}
\caption{\textbf{Mean-centering outperforms a 4$\times$ larger dictionary.} Number of biological concepts (out of 187 SwissProt concepts) captured by at least one SAE feature with $F_1 > 0.7$, on ESM3 representative middle layer, $k = 16$. A mean-centered 2048-feature SAE recovers more concepts than a baseline 8192-feature SAE.}
\label{tab:concepts_by_dict}
\end{table}

Surviving features are also more monosemantic. We computed Monosemanticity Scores \citep{pach2025sparse} on high-death models from two domains: DINOv3 (vision, with CLIP-ViT as the independent embedding model) and ESM3 (protein, with ESM2). The metric measures semantic coherence of each feature's top-activating inputs. On ESM3, mean-centered SAEs produce more monosemantic features across the score distribution (e.g., 2$\times$ more features above MS${>}0.5$ at the same dictionary size; \Cref{fig:ms_score}), while raw neurons score near zero (confirming SAEs perform real decomposition). On DINOv3, baseline features have almost no meaningful monosemanticity while mean-centered features hold nonzero scores across most of the dictionary. Across our highest-death settings (synthetic activations, protein models, and vision models), all three evaluations show mean-centered features are higher quality.

\subsection{When mean-centering isn't enough}
\label{sec:mc_exceptions}

Mean-centering addresses outlier-induced death, but this does not account for all death: other geometric causes appear in some layers, and features can still die over the course of training depending on hyperparameters.

Mean-centering eliminates outlier-induced death at the representative layer for every model in our suite except one. The exception is Evo1, where 73\% of features remain dead at initialization and ${\sim}$58\% after training. Why does the fix that works everywhere else fail here? Evo1's activations are extremely low-rank: just 4 of 4096 principal components capture 99\% of the variance. When variance concentrates in so few directions, only the features aligned with those few PCs can win the TopK competition, and the rest never fire. Centering removes the mean offset but leaves this directional imbalance intact. The same low-rank pattern, less severe, appears in the early layers (L1--L2) of DINOv3-7B, ESM3, ESM2-3B, and ProGen2-base; their middle layers (our typical SAE training sites) are unaffected, which is why mean-centering suffices there. Mean-centering always helps, but its \emph{sufficiency} depends on whether the post-centering covariance is rich enough for diverse features to compete.

PCA whitening modifies the activations to equalize variance across principal components, cleanly eliminating Evo1's residual death and reaching 0\% on every other affected layer. Active Subspace Initialization \citep{wang2025attentionlayersaddlowdimensional} instead initializes features in the data's high-variance directions without modifying the activations; on our pathologically low-rank cases it requires more aggressive settings than Wang et al.\ tested, leaving its effectiveness less clear. Together, $\gamma$ and an effective-rank measurement are enough to choose preprocessing per layer: MC when $\gamma$ is high, PCA or ASI when the post-centering rank is low. See \Aref{sec:app_spectral} for full analysis of the low-rank death mechanism, the per-layer diagnostic, and the comparison of fixes.

Additionally, large learning rates or sparsity penalties can kill features during training, even from a death-free initialization (\Aref{sec:d_extended_res}).

\section{Related Work}
\label{sec:related-work}

\paragraph{Activation preprocessing for SAE training.}

Mean-centering has appeared in prior SAE work, but its usage has been inconsistent across the literature, with no clear account of when it matters or why. \citet{bricken2023} initialized the bias to the activation mean (equivalent to centering), but later Anthropic releases omit this step \citep{conerly2024}. \citet{gao2024} use centering, noting only that it ``helps training'' without ablation. \citet{conerly2025jumprelu_ant} initialize encoder biases so each feature fires at a target rate, a related technique that implicitly accounts for the mean. $\gamma$ predicts when centering is necessary, and the recovery analysis (Section~\ref{sec:how-recovery-works}) explains why training alone cannot compensate in time.

Separately, \citet{saraswatula_data_2025} show that PCA whitening improves SAE feature quality on SAEBench \citep{karvonen2025saebench} by making the optimization landscape more convex. We use whitening for a complementary reason: to revive dead features in low-rank layers.

\paragraph{Origins of dimension-level outliers.}

The outliers we study are an instance of a well-documented phenomenon: large transformers develop a small number of hidden dimensions with activation magnitudes far exceeding the rest \citep{dettmers2022llmint8}. Their causes have been traced to optimizer and architectural choices \citep{elhage2023privileged, outlierfeatures2024neurips}, and other work proposes architectural mitigations \citep{bondarenko2023quantizable, hu2024outeffhop, luo2025germ}. These works address a different type of outlier (per-token spikes, captured by their kurtosis and infinity norm metrics) and propose changes applied during pretraining rather than handling outliers in existing models. Recent work has found that both dimension level outlier activations and attention sinks function as important rescaling factors which stabilize training \citep{qiu2026unified}.

\section{Discussion}

Feature death in SAEs is often not a training problem at all. In models with high outlier severity ($\gamma$), activation geometry determines which features will live and die before the first gradient step. The two death pathways (dead-by-ReLU and dead-by-TopK) originate from the same geometric cause but recover on very different timescales. Bias learning is the bottleneck at high $\gamma$, explaining why AuxK rescues some models but not others.

We did not initially expect the mechanism to be this direct. A more natural hypothesis (and the one we started with) was that outliers corrupt gradients or distort the loss landscape, causing training to slowly break down. Instead, the damage is done at initialization, and no amount of training recovers it within practical compute budgets.

Reviving dead features yields better SAEs at the same size: mean-centered SAEs match a 4$\times$ larger baseline on concept recovery and are more monosemantic (\Cref{fig:ms_score}).

\paragraph{Practical guidance.} For most models, initialize the SAE bias to the geometric median of activations: this eliminates outlier-induced death at zero runtime cost. The exception is models with intrinsically low-rank activations, where a few principal components carry nearly all the variance. The post-LN effective rank of the activation covariance, computable on a single batch, predicts when this occurs: below ${\sim}2\%$ of hidden dimension, MC alone leaves substantial residual death and PCA whitening reaches 0\% in every case (\Cref{tab:spectral_fixes}). Active Subspace Initialization \citep{wang2025attentionlayersaddlowdimensional} is an alternative that modifies the encoder initialization rather than the activations; it works well in moderately low-rank settings but has unclear effectiveness in the more extreme cases we see.

Variable death rates also affect cross-layer methods that chain feature dictionaries across layers, such as transcoders \citep{dunefsky2024transcoders}; mean-centering stabilizes effective capacity, which may help.

\paragraph{Limitations.}
Our analysis focuses on TopK SAEs and residual stream activations; ReLU and JumpReLU SAEs also exhibit outlier-induced death but experience additional training-time death from sparsity pressure that we don't analyze. $\gamma$ explains much of the variation in dead-feature rates at initialization, but other geometric properties also contribute: heavy tails reduce death below the formula's prediction (\Aref{sec:heavy_tails}), and low-rank structure causes additional death beyond what $\gamma$ captures (\Aref{sec:app_spectral}).

\paragraph{Open questions.}
Why outlier severity differs across architectures and domains is an open practical question. A similar open thread is LayerNorm, another common SAE preprocessing step typically motivated by easier hyperparameter transfer \citep{templeton2024scaling}. We find that LN changes training dynamics but neither resolves outlier-induced death nor matches mean-centering on feature recovery (\Cref{fig:centering_synthetic}, \Aref{sec:d_extended_res}). How LN affects dynamics across more models, particularly those with large per-token scale variation, is worth further study.

\section*{Acknowledgments}

We thank Mert Yuksekgonul for helpful feedback on the first draft of this paper, and members of the Zou lab for valuable conversations. E.S. is supported by NSF GRFP (grant no. DGE-2146755). E.A. is supported by NIH T32 training grant (grant no. 1T32GM158494-01).

\section*{Impact Statement}

This paper presents work whose goal is to advance the field of Machine Learning, specifically mechanistic interpretability. By improving our ability to understand the internal representations of neural networks across domains (language, vision, biology), this work may contribute to AI safety and alignment research, and improve our ability to study diverse model types including frontier biological models which can provide scientific insights through analysis.

\bibliography{references}
\bibliographystyle{icml2026}

\newpage
\beginappendix

\setcounter{figure}{0}
\renewcommand{\thefigure}{S\arabic{figure}}
\setcounter{table}{0}
\renewcommand{\thetable}{S\arabic{table}}

\printappendixtoc
\clearpage

\section{Experimental Setup}
\label{app:exp_setup}

\subsection{Datasets}

\subsubsection{Real-world activation datasets}

We extract activations from a range of pretrained models spanning
language, vision, protein, and genomic modalities. Per-model details are
in \Cref{tab:model-dataset-layer}.

\begin{table}[h]
\centering
\small
\begin{tabular}{l r r l r}
\hline
Model Name & Rep.\ Layer & \# Layers Swept & Dataset & \# Tokens \\
\hline
GPT-2 \cite{radford2019} & 6 & 11 & OpenWebText \cite{gokaslan2019openwebtext} & 10{,}000{,}000 \\
Pythia-410M \cite{biderman2023} & 12 & 23 & OpenWebText & 10{,}000{,}000 \\
Pythia-70M \cite{biderman2023} & --- & 5 & OpenWebText & 10{,}000{,}000 \\
ModernBERT-Large \cite{warner2025smarter} & 14 & 27 & OpenWebText & 10{,}000{,}000 \\
ModernBERT \cite{warner2025smarter} & --- & 21 & OpenWebText & 10{,}000{,}000 \\
DINOv2-L \cite{oquab2024} & 12 & 23 & CIFAR-10 \cite{krizhevsky2009learning} & 10{,}000{,}000 \\
DINOv2-B \cite{oquab2024} & --- & 11 & CIFAR-10 & 10{,}000{,}000 \\
DINOv3-7B \cite{simeoni2025dinov3} & 20 & 39 & CIFAR-10 & 10{,}000{,}000 \\
DINOv3-B \cite{simeoni2025dinov3} & 6 & 11 & CIFAR-10 & 10{,}000{,}000 \\
Stable Diffusion 3.5 Large \cite{esser2024scalingrectifiedflowtransformers} & 37 & 37 & CIFAR-10 & 10{,}000{,}000 \\
ESM2-3B \cite{lin2023} & 18 & 35 & Swiss-Prot \cite{uniprot2025uniprot} & 10{,}000{,}000 \\
ESM2-650M \cite{lin2023} & 16 & 32 & Swiss-Prot & 10{,}000{,}000 \\
ESM2-35M \cite{lin2023} & --- & 11 & Swiss-Prot & 10{,}000{,}000 \\
ESM3 \cite{hayes2025simulating} & 24 & 47 & Swiss-Prot & 10{,}000{,}000 \\
ProGen2-large \cite{nijkamp2023progen2} & 15 & 31 & Swiss-Prot & 10{,}000{,}000 \\
ProGen2-base \cite{nijkamp2023progen2} & --- & 26 & Swiss-Prot & 10{,}000{,}000 \\
gLM2 \cite{cornman2024omg} & 15 & 32 & Swiss-Prot & 10{,}000{,}000 \\
AlphaFold3 \cite{abramson2024} & Pairformer single rep (no recycle) & 1 & UniRef \cite{suzek2015uniref} & 10{,}000{,}000 \\
Evo1 \cite{nguyen2024sequence} & 14 & 31 & EMBL European Nucleotide Archive \cite{david2026european} & 10{,}000{,}000 \\
\hline
\multicolumn{2}{r}{\textbf{Total}} & \textbf{454} & & \\
\hline
\end{tabular}
\caption{Models, datasets, and per-model layer counts. \textit{Rep.\ Layer} gives the representative middle-of-network layer used in main figures (e.g.\ Figures~\ref{fig:death_patterns},~\ref{fig:centering_synthetic},~\ref{fig:extended_all_models}); a dash indicates the model is included only in the cross-model $\gamma$-vs-death analysis (Figure~\ref{fig:gamma_combined}) and has no rep-layer experiments. \textit{\# Layers Swept} gives the number of layers each model contributes to the cross-model sweep, excluding the embedding layer (layer 0). All extractions use 10M tokens.}
\label{tab:model-dataset-layer}
\end{table}

\begin{figure}[p]
\centering
\includegraphics[width=0.95\textwidth]{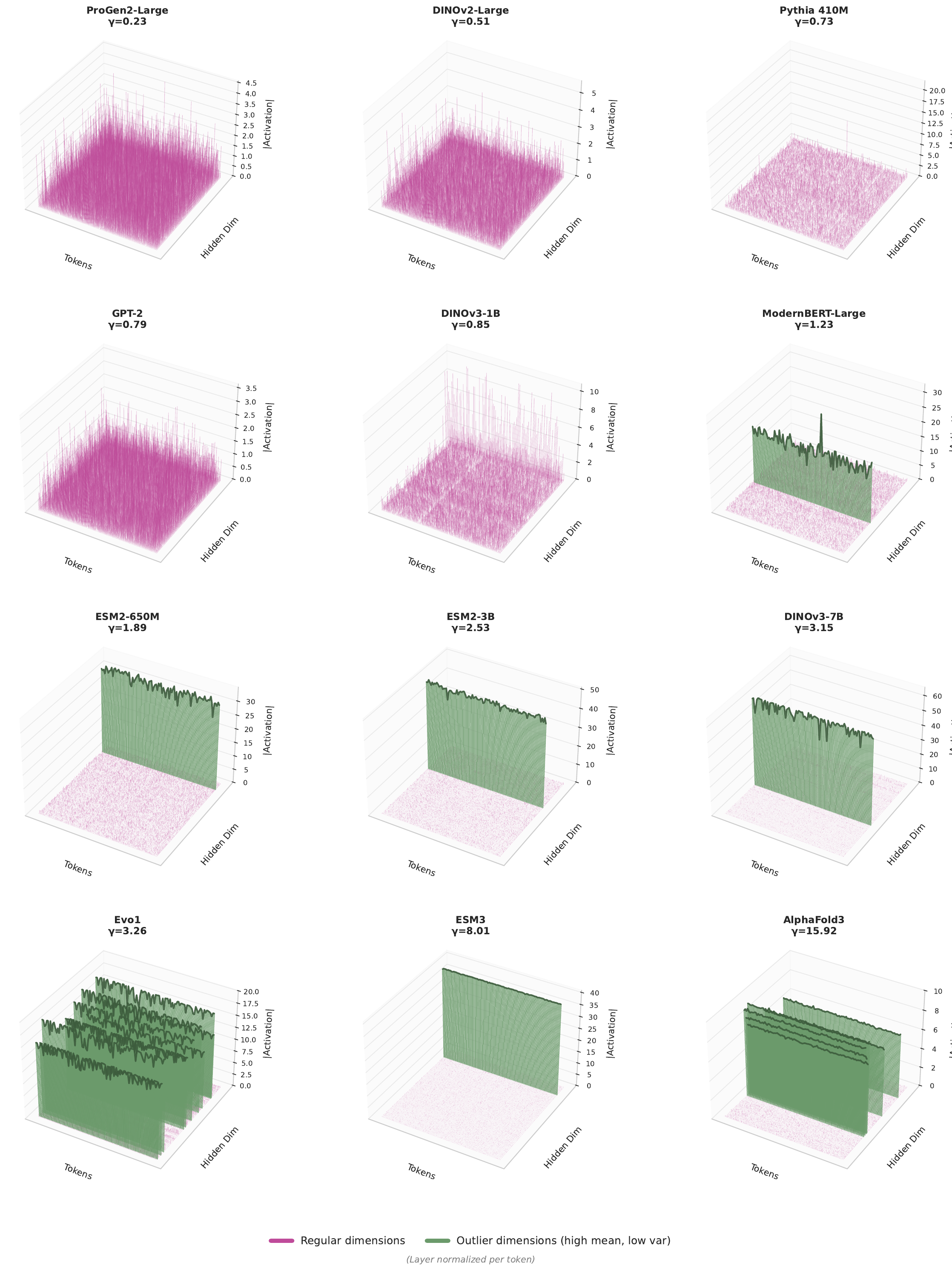}
\caption{\textbf{Dimension-level outliers appear as persistent
high-magnitude ridges across tokens.} Activation magnitude for each model at
its representative layer, ordered by increasing $\gamma$. Green: dimensions
with high mean and low per-token variance. LayerNorm-normalized per token.}
\label{app:fig_skyline_acts_all_models}
\end{figure}

\subsubsection{Synthetic activation datasets}
\label{app:synthetic_data_details}

We generate synthetic data following a sparse linear model. Each sample
$\mathbf{x} \in \mathbb{R}^{50}$ is generated as:
\begin{equation}
    \mathbf{x} = \mathbf{F}^\top \mathbf{c} + \boldsymbol{\mu}_{\text{outlier}}
\end{equation}
where $\mathbf{F} \in \mathbb{R}^{100 \times 50}$ contains 100 ground
truth feature directions (random unit vectors), $\mathbf{c} \in
\mathbb{R}^{100}$ is a sparse activation vector, and
$\boldsymbol{\mu}_{\text{outlier}} \in \mathbb{R}^{50}$ is a constant
bias vector.

For each sample, exactly 5 features are active (5\% sparsity), selected
uniformly at random. Active coefficients are drawn from $c_i \sim
|\mathcal{N}(0,1)| + 0.5$ (shifted half-normal), ensuring positive
activations with minimum 0.5.

The outlier bias vector has a single non-zero entry,
$(\boldsymbol{\mu}_{\text{outlier}})_0$, set to a configurable value
while the remaining entries are zero. Varying this value produces
synthetic data spanning approximately $\gamma \in [1, 50]$, where
$\gamma = \|\boldsymbol{\mu}\| / \|\boldsymbol{\sigma}\|$ as defined in
the main text; the specific $\gamma$ values used are visible as data
points in \Cref{fig:gamma_combined}.

By default, each synthetic SAE uses dictionary size 100 (matching the
number of ground truth features), TopK with $k=5$ (matching the number
of active features per sample), batch size 4096, and trains for 1--2M
steps on 1M samples (specific durations are listed with figures). These
defaults are varied in ablations in \Aref{sec:gamma_width_k}.

\begin{table}[h]
\centering
\small
\begin{tabular}{ll}
\toprule
\textbf{Parameter} & \textbf{Default value} \\
\midrule
Data dim ($d$) & 50 \\
Ground truth features ($m$) & 100 \\
Sparsity & 0.05 (5 active per sample) \\
Dataset size & 1{,}000{,}000 samples \\
\midrule
Dictionary size & 100 \\
TopK $k$ & 5 \\
Batch size & 4096 \\
Training steps & 1--2{,}000{,}000 \\
\bottomrule
\end{tabular}
\caption{Synthetic data parameters.}
\label{tab:synthetic_params}
\end{table}

\subsection{SAE Architectures and Training}

We use ReLU SAEs as described in \citep{conerly2024}, TopK SAEs as
described in \citep{gao2024}, and Jump-ReLU SAEs as described in
\citep{rajamanoharan2024jumpingaheadimprovingreconstruction}.

For SAEs trained on real-world activations, unless otherwise noted we
use a batch size of 4096 tokens, dictionary size 8192, and 100{,}000
training steps (${\approx}410$M tokens per training run). We sweep over
learning rates in the range $10^{-5}$ to $10^{-3}$ and select the run with
the lowest MSE. For SAEs trained on synthetic
activations, training configuration is given in
\Aref{app:synthetic_data_details} (\Cref{tab:synthetic_params}).

TopK SAE \cite{gao2024} hyperparameters:

\begin{table}[h]
\centering
\begin{tabular}{l r l}
\hline
Hyperparameter & Default & Description \\
\hline
TopK ($k$) & 64 & Number of active features selected per example \\
AuxK coefficient & 1/32 & Weight on the auxiliary TopK loss term (AuxK) \\
Examples until dead & 256{,}000 & Number of examples with no activation after which a unit is considered ``dead'' \\
\hline
\end{tabular}
\caption{TopK hyperparameters.}
\label{tab:topk-hparams}
\end{table}

JumpReLU SAE \cite{rajamanoharan2024jumpingaheadimprovingreconstruction} hyperparameters:

\begin{table}[H]
\centering
\begin{tabular}{l r l}
\hline
Hyperparameter & Value & Description \\
\hline
$\theta$ & 0.001 & Initial JumpReLU threshold value \\
 $\epsilon$ & 0.001 & Straight-through estimator (STE) bandwidth \\
L0 penalty weight & 1 & L0 sparsity penalty weight (see Eq. 16 in \citet{rajamanoharan2024jumpingaheadimprovingreconstruction}) \\
target L0 & 64 & Targeted sparsity \\
\hline
\end{tabular}
\caption{JumpReLU hyperparameters.}
\label{tab:jumprelu-hparams}
\end{table}

ReLU SAE \cite{huben2024sparse} hyperparameters:

\begin{table}[H]
\centering
\begin{tabular}{l l p{7.5cm}}
\hline
Hyperparameter & Values & Description \\
\hline
L1 coefficient & $\{10^{-4},\,10^{-3},\,10^{-2},\,10^{-1}\}$ & Weight on the L1 sparsity penalty \\
Ghost gradients & \{True, False\} & Whether to use ghost gradients to provide learning signal for features that are inactive (helps revive/learn rarely-active or dead features). \\
\hline
\end{tabular}
\caption{ReLU hyperparameters.}
\label{tab:relu-hparams}
\end{table}

\subsection{Evaluation Metrics}
\label{app:metrics}

We evaluate SAEs on the following metrics, applied as appropriate to
each experimental setting:

\begin{itemize}
    \item \textbf{Dead feature percentage}: Fraction of latents that
    never activate on 100k held-out examples. During training, we
    consider features dead if they have not activated for 256k examples.

    \item \textbf{MSE}: Mean squared reconstruction error
    $\|\mathbf{x} - \hat{\mathbf{x}}\|^2$.

    \item \textbf{MMCS (synthetic only)}: Mean max cosine similarity
    with ground-truth features. For each ground-truth feature direction,
    we compute its maximum cosine similarity with any learned dictionary
    direction, then average these maxima across ground-truth features.
    MMCS is high when every ground-truth feature has at least one
    learned feature pointing in (approximately) the same direction.

    \item \textbf{Biological concept coverage}: For each of 187 SwissProt
    biological concepts \citep{uniprot2025uniprot}, we train a linear
    probe per SAE feature predicting that concept on 10{,}000 random
    SwissProt sequences, following \citet{simon2025interplm}. We
    evaluate at ESM3 layer 24. We report the number of concepts that
    have at least one SAE feature achieving $F_1 > 0.7$, as a coverage
    metric for biologically meaningful structure. To avoid evaluating concepts unlikely to have associated features, we selected 187 concepts (out of a total of 833 concepts) that had previously shown at least moderate performance in ESM2, defined as an F1 score greater than 0.3.

    \item \textbf{Monosemanticity Score (MS)}: Defined by
    \citet{pach2025sparse} as the semantic coherence of each feature's
    top-activating inputs under an independent embedding model. We use
    CLIP-ViT for DINOv3 features and ESM2 for ESM3 features. Higher MS
    indicates a feature's top activations cluster tightly in the
    independent model's embedding space, suggesting the feature
    represents a single meaningful concept.
\end{itemize}

\begin{figure}[H]
\centering
\includegraphics[width=0.95\textwidth]{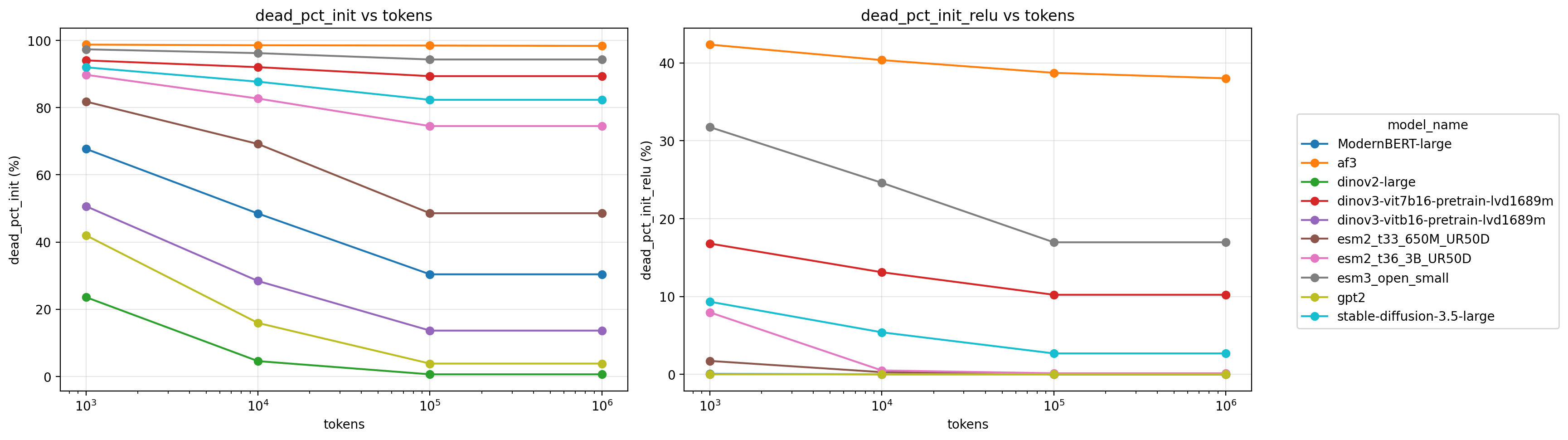}
\caption{\textbf{Dead feature rates plateau around 100,000 evaluation
examples.} Left: total dead features; right: dead-by-ReLU. Both stabilize by
$10^5$ examples across modalities.}
\label{fig:dead_feat_thresholds_for_eval}
\end{figure}

The 100k held-out threshold for the dead-feature metric is empirically
justified: dead-feature rates plateau by ${\sim}10^5$ examples across
modalities (\Cref{fig:dead_feat_thresholds_for_eval}).

\section{Why $\gamma$ predicts dead features: derivation and validation}
\label{sec:derivation}

We want to predict, at random initialization, how many SAE features die.
A feature dies when its pre-activation never crosses some threshold
across the evaluation set: zero for ReLU, the input-dependent rank-$k$
boundary for TopK. We show below that both reduce, to a good
approximation, to a closed-form expression in a single scalar quantity:
$\gamma = \|\bm{\mu}\| / \|\bm{\sigma}\|$, the activation mean's
norm divided by the per-dimension standard deviation's norm. The
dead-by-ReLU rate is $\Phi(-C/\gamma)$ and the dead-by-TopK rate is
$\Phi(t_k - C/\gamma)$ (Section~\ref{sec:analytical}).

\paragraph{Approach.} The derivation has four steps:
\begin{enumerate}
    \item Each feature's encoder weight $\mathbf{w}_i$ is a random unit
    vector. Its projection onto any fixed vector is approximately
    Gaussian (\Aref{sec:projection_gaussian}).
    \item Use this to decompose the pre-activation
    $z_i(\mathbf{x}) = s_i + r_i(\mathbf{x})$ into a data-independent
    \emph{shift} $s_i = \mathbf{w}_i \cdot \bm{\mu}$ and a
    data-dependent \emph{signal} $r_i(\mathbf{x}) = \mathbf{w}_i \cdot
    (\mathbf{x} - \bm{\mu})$, both with Gaussian distributions
    parameterized by $\bm{\mu}$ and $\bm{\sigma}$.
    \item For each death pathway (dead-by-ReLU and dead-by-TopK), find
    the shift threshold below which no sample can rescue the feature.
    \item Compute the fraction of features whose shift falls below this
    threshold; the dead-by-TopK extension follows the same approach
    with a different threshold.
\end{enumerate}

\paragraph{Assumptions.}
The key assumption: for a fixed encoder weight $\mathbf{w}_i$, the
signal $\mathbf{w}_i \cdot (\mathbf{x} - \bm{\mu})$ is approximately
Gaussian over the data distribution. This holds when activations are
not too heavy-tailed; heavier-tailed activations rescue features the
formula would predict dead (\Aref{sec:heavy_tails} quantifies when).
We also assume evaluation samples are approximately independent.

\paragraph{Measurement note.} When computing $\gamma$ on real data, we
apply per-token LayerNorm before measuring $\bm{\mu}$ and
$\bm{\sigma}$. This removes between-token scale variation that would
otherwise inflate $\|\bm{\sigma}\|$ and understate the outlier
severity (\Aref{sec:per_token_scale}).

\subsection{Projections of Random Unit Vectors Are Approximately Gaussian}
\label{sec:projection_gaussian}

Encoder weights are initialized by sampling i.i.d.\ entries from
$\mathcal{N}(0, 1)$ and normalizing each row to unit norm, which is
equivalent to sampling uniformly from the unit sphere in $d$ dimensions.

The projection of such a random unit vector $\mathbf{w}$ onto any fixed
vector $\bm{\mu}$ is approximately Gaussian with mean zero and variance
$\|\bm{\mu}\|^2 / d$. Write $\mathbf{w} = \mathbf{g}/\|\mathbf{g}\|$
where $\mathbf{g} \sim \mathcal{N}(0, \mathbf{I}_d)$. The unnormalized
projection $\mathbf{g} \cdot \bm{\mu}$ is exactly $\mathcal{N}(0,
\|\bm{\mu}\|^2)$, since it is a weighted sum of independent Gaussians.
In high dimensions, $\|\mathbf{g}\|$ concentrates around $\sqrt{d}$, so
dividing by $\|\mathbf{g}\|$ scales the variance by $1/d$:
\begin{equation}
\mathbf{w} \cdot \bm{\mu} \;\sim\; \mathcal{N}\!\left(0,\; \frac{\|\bm{\mu}\|^2}{d}\right)
\end{equation}

\subsection{Pre-activations Decompose into a Fixed Shift and a Varying Signal}

The pre-activation for feature $i$ on input $\mathbf{x}$ decomposes as:
\begin{equation}
z_i(\mathbf{x}) = \mathbf{w}_i \cdot \mathbf{x} = \underbrace{\mathbf{w}_i \cdot \bm{\mu}}_{s_i \text{ (shift)}} + \underbrace{\mathbf{w}_i \cdot (\mathbf{x} - \bm{\mu})}_{r_i(\mathbf{x}) \text{ (signal)}}
\end{equation}

The shift $s_i$ is locked in once encoder weights are drawn; it is identical for every input the SAE will ever process. The signal $r_i(\mathbf{x})$ is the part that responds to what the input contains. If the shift dominates, the feature fires (or doesn't) regardless of input.

\paragraph{Shift distribution.} The shift $s_i = \mathbf{w}_i \cdot \bm{\mu}$ is a projection of a random unit vector onto the fixed mean $\bm{\mu}$. Applying the result from the previous subsection:
\begin{equation}
s_i \sim \mathcal{N}\!\left(0,\; \frac{\|\bm{\mu}\|^2}{d}\right)
\end{equation}

\paragraph{Signal distribution.} The signal $r_i(\mathbf{x}) = \mathbf{w}_i \cdot (\mathbf{x} - \bm{\mu})$ varies with each input. For a fixed input, the same projection argument gives variance $\|\mathbf{x} - \bm{\mu}\|^2/d$, which differs across inputs. To characterize typical signal magnitude:
\begin{align}
\mathbb{E}_\mathbf{x}\!\left[\|\mathbf{x} - \bm{\mu}\|^2\right] &= \sum_{j=1}^d \text{Var}(x_j) = \sum_{j=1}^d \sigma_j^2 = \|\bm{\sigma}\|^2
\end{align}
When $d$ is large and dimensions are not strongly correlated, this sum of $d$ squared deviations concentrates tightly around its expectation, with relative fluctuation shrinking as $1/\sqrt{d}$. So for a typical input:
\begin{equation}
r_i(\mathbf{x}) \;\sim\; \mathcal{N}\!\left(0,\; \frac{\|\bm{\sigma}\|^2}{d}\right) \quad \text{(for typical inputs)}
\end{equation}

\subsection{Features Die When Their Shift Exceeds $C$ Standard Deviations}

A feature is dead-by-ReLU if $z_i(\mathbf{x}) = s_i + r_i(\mathbf{x}) < 0$ for all $N$ evaluation samples. Each sample is one chance for the signal to rescue the feature: survival requires just a single input where the signal overcomes the negative shift. The question is how negative the shift must be for none of $N$ samples to provide enough signal.

Let $\sigma_r = \|\bm{\sigma}\|/\sqrt{d}$ denote the typical signal standard deviation. For a feature with shift $s_i = -t$ (where $t > 0$), the probability that a single sample's signal exceeds $t$ is $P(r_i(\mathbf{x}) > t) = 1 - \Phi(t/\sigma_r)$. Across $N$ independent samples, the expected number that exceed $t$ is $N \cdot [1 - \Phi(t/\sigma_r)]$.

We define the lethal threshold as the shift where this expected count equals 1, below which fewer than one sample is expected to rescue the feature. The exact cutoff matters less than it might seem: $C$ varies only from 3.7 to 4.8 as $N$ ranges from $10^4$ to $10^6$, so predicted death rates are insensitive to the precise choice. Setting the expected count to 1 and solving:
\begin{align}
N \cdot \left[1 - \Phi\!\left(\frac{t}{\sigma_r}\right)\right] &= 1 \\[4pt]
\Phi\!\left(\frac{t}{\sigma_r}\right) &= 1 - \frac{1}{N} \\[4pt]
\frac{t}{\sigma_r} &= \Phi^{-1}\!\left(1 - \frac{1}{N}\right) \;\equiv\; C
\end{align}

The lethal threshold is therefore $t^* = C \cdot \sigma_r = C \cdot \|\bm{\sigma}\|/\sqrt{d}$. Features with shift more negative than $-t^*$ would need a signal fluctuation so large that we expect fewer than one such event across all $N$ samples.

\paragraph{Values of $C$:}
\begin{center}
\begin{tabular}{@{}cc@{}}
\toprule
$N$ & $C = \Phi^{-1}(1 - 1/N)$ \\
\midrule
$10^3$ & 3.09 \\
$10^4$ & 3.72 \\
$10^5$ & 4.26 \\
$10^6$ & 4.75 \\
\bottomrule
\end{tabular}
\end{center}

For our evaluation with $N = 100{,}000$: $C = 4.26$.

\subsection{The Death Rate Follows From $\gamma$ in Closed Form}

All that remains is to ask what fraction of features drew a shift this unlucky. Since $s_i \sim \mathcal{N}(0, \|\bm{\mu}\|^2/d)$, the probability that a feature's shift falls below $-t^* = -C \cdot \|\bm{\sigma}\|/\sqrt{d}$ is:
\begin{align}
P(\text{dead-by-ReLU}) &= P\!\left(s_i < -C \cdot \frac{\|\bm{\sigma}\|}{\sqrt{d}}\right) \\[6pt]
&= P\!\left(\frac{s_i}{\|\bm{\mu}\|/\sqrt{d}} < \frac{-C \cdot \|\bm{\sigma}\|/\sqrt{d}}{\|\bm{\mu}\|/\sqrt{d}}\right) \\[6pt]
&= P\!\left(Z < \frac{-C \cdot \|\bm{\sigma}\|}{\|\bm{\mu}\|}\right) \quad \text{where } Z \sim \mathcal{N}(0,1) \\[6pt]
&= \Phi\!\left(\frac{-C}{\gamma}\right)
\end{align}
where $\gamma = \|\bm{\mu}\|/\|\bm{\sigma}\|$. The embedding dimension $d$ drops out entirely: both the shift variance ($\|\bm{\mu}\|^2/d$) and the signal variance ($\|\bm{\sigma}\|^2/d$) scale as $1/d$, so their ratio is dimension-free. This is why $\gamma$ works as a diagnostic across models with embedding dimensions ranging from 384 (AlphaFold3) to 4096 (DINOv3, Evo1).

The final result:
\begin{equation}
\boxed{P(\text{dead-by-ReLU}) = \Phi\!\left(\frac{-C}{\gamma}\right), \quad C = \Phi^{-1}(1 - 1/N)}
\end{equation}

\paragraph{Limiting behavior.} As $\gamma \to 0$ (no outliers), the argument of $\Phi$ goes to $-\infty$ and $P(\text{dead}) \to 0$: without outliers, random features are not born dead. As $\gamma \to \infty$ (extreme outliers), the argument approaches $0$ and $P(\text{dead}) \to 0.5$: feature fate is determined entirely by the sign of $\mathbf{w}_i \cdot \bm{\mu}$, with positive and negative alignment equally likely under symmetric initialization.

\paragraph{Multiple outlier dimensions.} The formula does not assume outliers are concentrated in a single dimension. The shift $s_i = \mathbf{w}_i \cdot \bm{\mu}$ and its variance $\|\bm{\mu}\|^2/d$ depend only on the norm of $\bm{\mu}$, not on how it is distributed across coordinates. For example, one dimension with a mean of 1000 and ten dimensions each with a mean of 316 produce the same $\gamma$ and the same predicted death rate.

\subsection{Extension to Dead-by-TopK}
\label{sec:topk_derivation}

The dead-by-ReLU derivation asks whether a feature's shift is so negative
that no input can push its pre-activation above zero. Dead-by-TopK raises
the bar: a feature must not merely be positive, but rank among the top $k$
out of $n$ features on at least one input.

\paragraph{The threshold is approximately fixed across inputs at high $\gamma$.}
At high $\gamma$, the shift $s_i = \mathbf{w}_i \cdot \bm{\mu}$ dominates
every feature's pre-activation: shifts are spread across features by a
factor $\gamma$ more than the signal $r_i(\mathbf{x})$ can move any single
feature. The signal therefore cannot reorder features whose shifts already
differ by several times its range, and the top-$k$ selection picks
essentially the same $k$ features on every input: the $k$ with the
largest shifts. The bar a feature must clear is therefore approximately
constant across inputs: it is the $(k+1)$-th largest shift across
features, equivalently the $(1 - k/n)$ quantile of the shift distribution.

Since shifts are i.i.d.\ draws from $\mathcal{N}(0, \|\bm{\mu}\|^2/d)$,
this quantile is:
\begin{equation}
\tau_k \;=\; \Phi^{-1}\!\left(1 - \frac{k}{n}\right) \cdot
\frac{\|\bm{\mu}\|}{\sqrt{d}}
\;=\; t_k \cdot \frac{\|\bm{\mu}\|}{\sqrt{d}},
\quad t_k \equiv \Phi^{-1}\!\left(1 - \frac{k}{n}\right)
\end{equation}
For $k = 64$ and $n = 8192$: $t_k = \Phi^{-1}(0.9922) = 2.42$.

\paragraph{Death criterion.}
A feature is dead-by-TopK if its pre-activation never exceeds $\tau_k$ on
any of $N$ evaluation samples:
\begin{equation}
\max_{j=1,\ldots,N}\; \bigl[s_i + r_i(\mathbf{x}_j)\bigr] \;<\; \tau_k
\end{equation}
Following the same argument as \Aref{sec:derivation}.3, the feature
survives only if some sample's signal exceeds the gap
$\tau_k - s_i$. The expected number of rescuing samples is
$N \cdot [1 - \Phi((\tau_k - s_i)/\sigma_r)]$, and setting this to 1
gives the lethal threshold: a feature dies when
\begin{equation}
s_i \;<\; \tau_k \;-\; C \cdot \sigma_r
\;=\; t_k \cdot \frac{\|\bm{\mu}\|}{\sqrt{d}}
\;-\; C \cdot \frac{\|\bm{\sigma}\|}{\sqrt{d}}
\end{equation}

\paragraph{Death rate.}
The fraction of features whose shift falls below this threshold is:
\begin{align}
P(\text{dead-by-TopK})
&= P\!\left(s_i < t_k \cdot \frac{\|\bm{\mu}\|}{\sqrt{d}}
   - C \cdot \frac{\|\bm{\sigma}\|}{\sqrt{d}}\right) \\[4pt]
&= \Phi\!\left(\frac{t_k \cdot \|\bm{\mu}\|/\sqrt{d}
   - C \cdot \|\bm{\sigma}\|/\sqrt{d}}{\|\bm{\mu}\|/\sqrt{d}}\right) \\[4pt]
&= \Phi\!\left(t_k - \frac{C}{\gamma}\right)
\end{align}
where $\gamma = \|\bm{\mu}\|/\|\bm{\sigma}\|$ as before. The dead-by-ReLU
formula $\Phi(-C/\gamma)$ is the special case $t_k = 0$: the ReLU threshold
is zero, so only the signal term matters.

\paragraph{Limiting behavior.}
As $\gamma \to \infty$, the argument of $\Phi$ approaches $t_k =
\Phi^{-1}(1 - k/n)$, giving $P(\text{dead}) \to 1 - k/n$. For $k = 64$,
$n = 8192$, this is $99.2\%$: only the $k$ features most positively aligned
with $\bm{\mu}$ survive, and the rest are dead regardless of input content.

\subsection{$\gamma$ predicts death across SAE width and sparsity $k$}
\label{sec:gamma_width_k}

\begin{figure}[H]
\centering
\includegraphics[width=\linewidth]{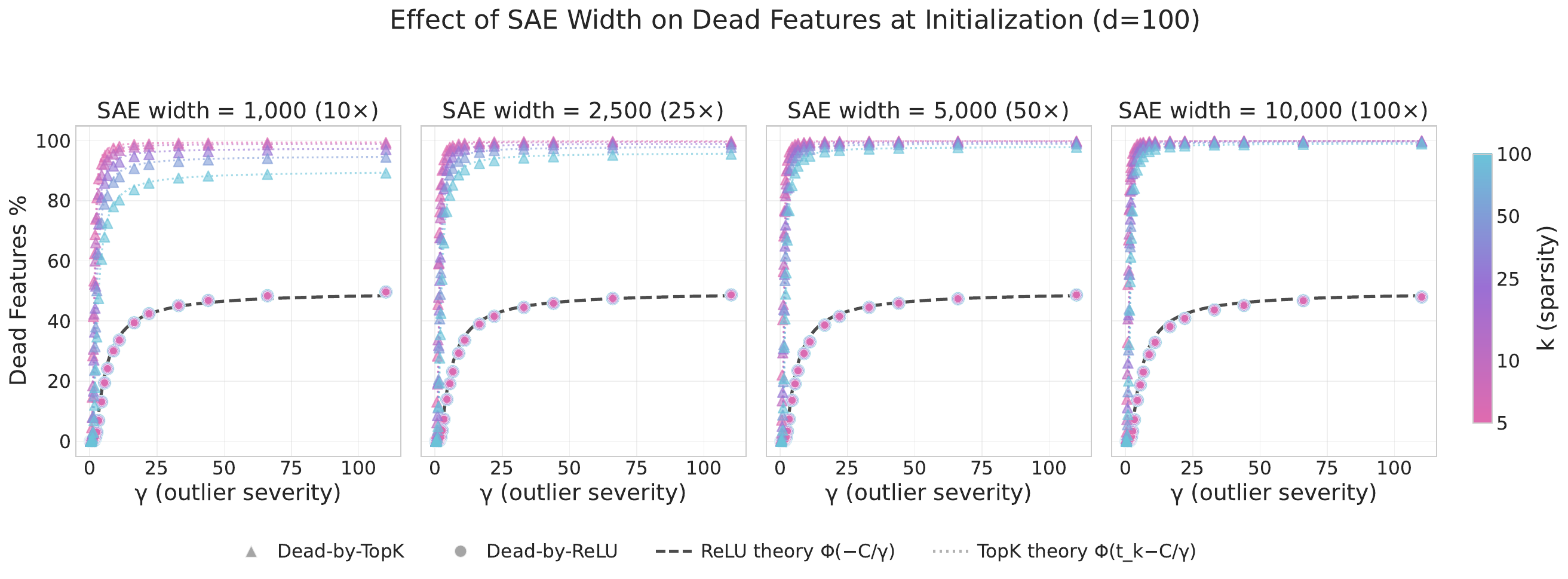}
\caption{\textbf{$\gamma$ predicts death rates across SAE width and sparsity.} Each panel shows a different SAE width (10$\times$ to 100$\times$ the input dimension); within each panel, color encodes sparsity $k \in \{5, 16, 32, 64, 100\}$ and marker shape distinguishes the two death pathways (triangles: dead-by-TopK; circles: dead-by-ReLU). Death rates rise monotonically with $\gamma$ on every (width, $k$) combination, confirming that the $\gamma$ diagnostic is robust to SAE hyperparameter choice. Dotted lines: per-$k$ dead-by-TopK theory $\Phi(t_k - C/\gamma)$. Dashed line: dead-by-ReLU theory $\Phi(-C/\gamma)$.}
\label{fig:gamma_width_k}
\end{figure}

The main-text analysis fixes the SAE width and sparsity for clarity. To check that $\gamma$ predicts death across these axes too, we trained TopK SAEs on synthetic activations spanning a range of $\gamma$ values, jointly sweeping SAE width $\in \{10\times, 25\times, 50\times, 100\times\}$ the input dimension and sparsity $k \in \{5, 16, 32, 64, 100\}$. \Cref{fig:gamma_width_k} shows the result.

\subsection{Death tracks $\gamma$ whether driven by $\boldsymbol{\mu}$ or $\boldsymbol{\sigma}$}
\label{sec:gamma_variance_intervention}

\begin{figure}[H]
\centering
\includegraphics[width=\linewidth]{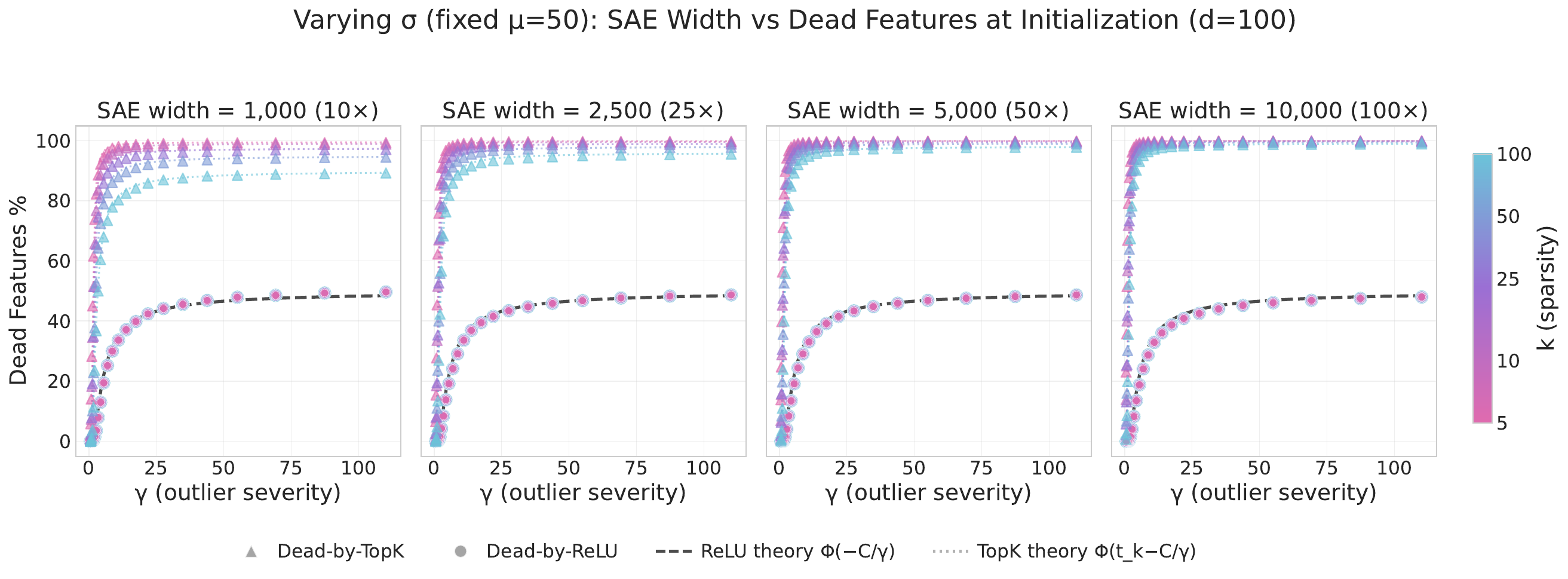}
\caption{\textbf{Death tracks $\gamma$ whether driven by $\boldsymbol{\mu}$ or $\boldsymbol{\sigma}$.} Same four-panel sweep over SAE width and sparsity $k$ as \Cref{fig:gamma_width_k}, but $\gamma$ is now varied by sweeping $\boldsymbol{\sigma}$ at fixed $\boldsymbol{\mu}=50$ rather than by sweeping the outlier mean. Death rates collapse onto the same $\gamma$-determined curves as in \Cref{fig:gamma_width_k}, confirming that $\gamma = \|\boldsymbol{\mu}\|/\|\boldsymbol{\sigma}\|$ is the operative quantity, not $\boldsymbol{\mu}$ or $\boldsymbol{\sigma}$ alone. Markers and theory lines as in \Cref{fig:gamma_width_k}.}
\label{fig:gamma_variance_intervention}
\end{figure}

The $\gamma$ formula combines $\|\boldsymbol{\mu}\|$ and $\|\boldsymbol{\sigma}\|$ into a single ratio. To test that the ratio (rather than either component alone) is the operative quantity, we ran a complementary sweep to \Cref{fig:gamma_width_k}: instead of varying $\gamma$ through the outlier mean, we fix $\boldsymbol{\mu}$ and sweep the per-dimension std $\boldsymbol{\sigma}$. Death rates collapse onto the same $\gamma$-determined curves (\Cref{fig:gamma_variance_intervention}), confirming that the ratio $\gamma$ itself, not $\boldsymbol{\mu}$ or $\boldsymbol{\sigma}$ alone, is what predicts death.

\subsection{Per-token scale variation affects $\gamma$ measurement}
    \label{sec:per_token_scale}

    $\gamma$ is computed from $\bm{\mu}$ and $\bm{\sigma}$ measured across
    tokens, but the structure that drives feature death is per-token: a feature
    is alive iff it ranks among the top $k$ pre-activations on at least one
    token. These rankings on any single token are unaffected by that token's
    overall norm, since rescaling a token multiplies every feature's
    pre-activation by the same positive factor and leaves their order
    unchanged. So feature death does not depend on how token norms vary across
    the dataset. Raw $\gamma$ does: tokens with large norms inflate
    $\bm{\sigma}$ more than $\bm{\mu}$ (variance squares deviations). When
    tokens have very different norms (typical of vision transformers), raw
    $\gamma$ understates outlier severity without the actual death rate
    changing.

    To measure the per-token directional structure that actually drives death,
    we normalize each token to unit norm (per-token LayerNorm) before computing
    $\bm{\mu}$ and $\bm{\sigma}$. SAEs are trained on the original unnormalized
    activations; we use the normalized activations only for the $\gamma$
    measurement. Per-token rescaling to a target norm (e.g.,
    $\sqrt{d_\text{model}}$) \citep{templeton2024scaling} has the same effect.

    The empirical effect is exactly what we'd expect: per-token LN changes
    $\gamma$ for models with high scale variation and leaves it largely
    unchanged for models with naturally uniform token norms
    (Table~\ref{tab:ln_dead_change}). Vision transformers exhibit the largest
    per-token scale variation: DINOv3-7B layers have $\gamma_\text{LN} /
    \gamma_\text{raw}$ ratios up to $16.9\times$ (median $5.3\times$). Without
    normalization, DINOv3-7B's within-model correlation between $\gamma$ and
    death is \emph{inverted} ($\rho = -0.61$), because the layers with the
    worst outliers also have the most per-token scale variation, masking the
    signal. Language models show moderate effects (GPT-2: median $2.5\times$).
    Protein language models are barely affected (ESM3: median $1.1\times$),
    because their tokens already have similar norms.

    \begin{table}[h]
    \centering
    \small
    \begin{tabular}{@{}llrc@{}}
    \toprule
    Model & Domain & $n_\text{layers}$ & $\gamma_\text{LN} / \gamma_\text{raw}$ \\
    \midrule
    DINOv3-7B & Vision & 40 & 5.3$\times$ (1.1--16.9) \\
    ModernBERT-Large & Language & 28 & 4.3$\times$ (1.0--10.7) \\
    GPT-2 & Language & 12 & 2.5$\times$ (1.0--4.3) \\
    ESM3 & Protein & 48 & 1.1$\times$ (1.0--1.3) \\
    AF3 & Protein & 1 & 1.1$\times$ \\
    \bottomrule
    \end{tabular}
    \caption{\textbf{Per-token LayerNorm changes the measured $\gamma$.}
    $\gamma_\text{LN} / \gamma_\text{raw}$: median ratio across layers (range in
    parentheses). Per-token normalization to a target norm (e.g.,
    $\sqrt{d_\text{model}}$) \citep{templeton2024scaling} has the same effect.}
    \label{tab:ln_dead_change}
    \end{table}
    
\subsection{Where the Formula Is Less Accurate: Heavy Tails Rescue Dead Features}
\label{sec:when_derivation_breaks}
\label{sec:heavy_tails}

In \Cref{fig:gamma_combined} (bottom panel), most dead-by-ReLU points (circles) track the theoretical curve closely, but some layers consistently fall below it: most visibly in ESM3 and ESM2-650M, where layers at $\gamma \approx 5$--$15$ have 5--15\% fewer dead-by-ReLU features than predicted. The cause is heavy-tailed signals. The derivation assumes the signal $r_i(\mathbf{x})$ is Gaussian, in which case its maximum across $N = 100{,}000$ samples is approximately $C \approx 4.26$ standard deviations of the signal distribution (i.e., $C \sigma_r$, where $\sigma_r$ is the per-feature signal std). A feature with shift $s_i$ below $-C \sigma_r$ cannot be rescued by any sample and is dead. When the signal is actually heavy-tailed, the maximum across $N$ samples can reach 8 or 9 $\sigma_r$; features with shifts down to $-8\sigma_r$ or $-9\sigma_r$ are then alive in reality but predicted dead by the formula.

\begin{figure}[H]
\centering
\includegraphics[width=\textwidth]{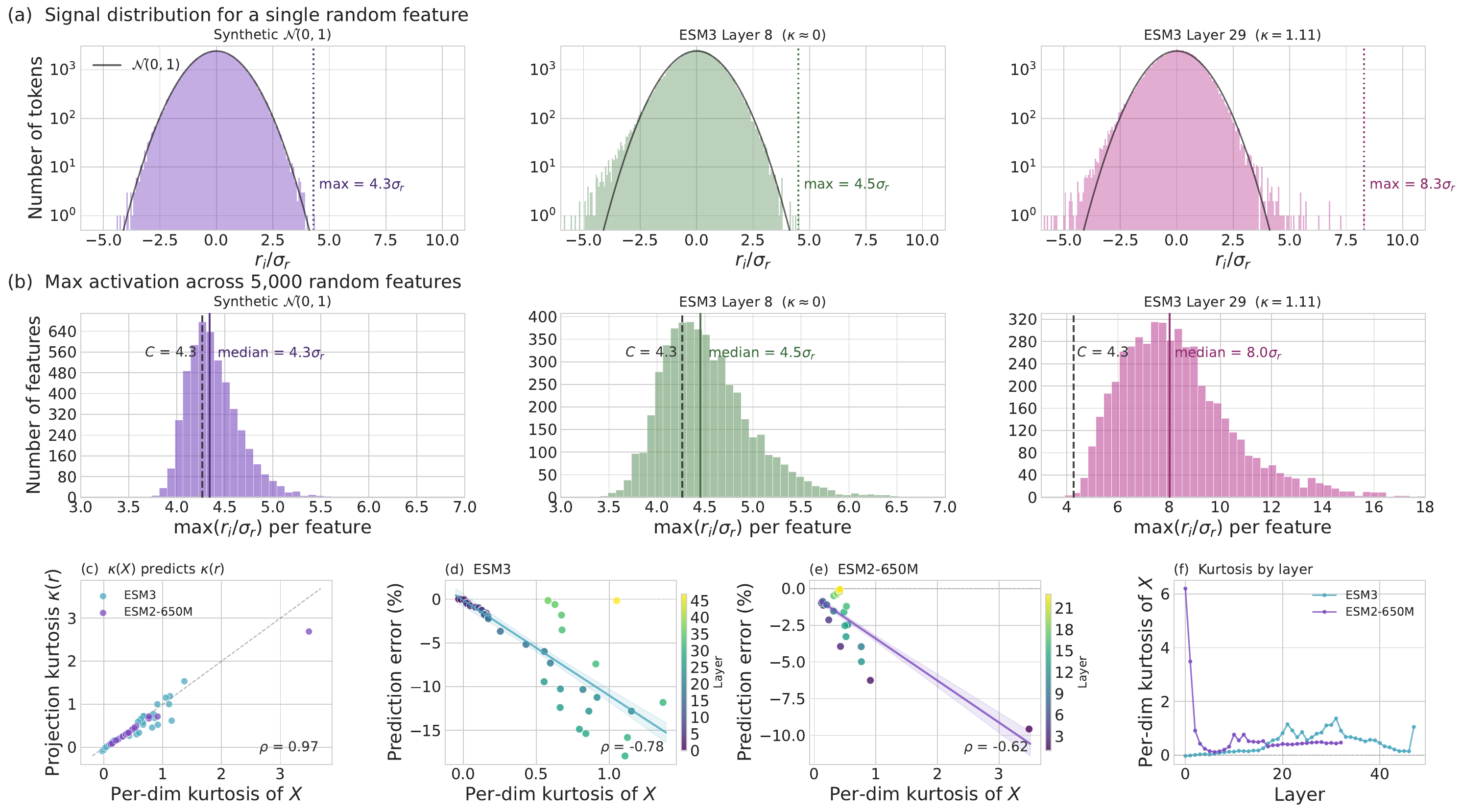}
\caption{Heavy-tailed embedding dimensions break the Gaussian assumption underlying the death formula. \textbf{(a)}~Signal distribution $r_i/\sigma_r$ for a single random feature direction, comparing synthetic Gaussian data, ESM3 layer~8 (near-Gaussian, $\kappa \approx 0$), and ESM3 layer~29 (heavy-tailed, $\kappa = 1.11$). The black curve shows the expected $\mathcal{N}(0,1)$ density; dotted lines mark the observed maximum. Layer~29's tails extend far beyond the Gaussian prediction. \textbf{(b)}~Distribution of $\max(r_i/\sigma_r)$ across 5{,}000 random feature directions. The dashed line marks the Gaussian-predicted maximum $C = 4.3$; the solid line marks the empirical median. For Gaussian and layer~8, the median is near $C$; for layer~29, the median is $8.0\sigma_r$, nearly double, rescuing features the formula predicts should be dead. \textbf{(c)}~Per-dimension kurtosis of $X$ predicts the kurtosis of random projections ($\rho = 0.97$), confirming the mechanistic link. \textbf{(d--e)}~Higher kurtosis correlates with larger formula overprediction (ESM3: $\rho = -0.78$; ESM2-650M: $\rho = -0.62$), colored by layer depth. \textbf{(f)}~Per-dimension kurtosis varies by layer, peaking in middle layers of ESM3 and early layers of ESM2-650M.}
\label{fig:kurtosis_mechanism}
\end{figure}

\Cref{fig:kurtosis_mechanism} illustrates this for ESM3. For a single random feature direction, the signal distribution on layer~8 ($\kappa \approx 0$) matches the Gaussian density and the observed maximum is $4.5\sigma_r$, close to the predicted $4.3\sigma_r$ (panel~a, middle). On layer~29 ($\kappa = 1.11$), the tails are visibly heavier and the maximum reaches $8.3\sigma_r$ (panel~a, right). This is not a fluke of one direction: across 5{,}000 random feature directions, the median maximum signal on layer~29 is $8.0\sigma_r$, nearly double the $4.3\sigma_r$ median for Gaussian data (panel~b). A feature with shift $s_i = -6\sigma_r$ would be permanently dead under Gaussian signals but alive on layer~29, because several samples exceed $+6\sigma_r$.

What makes some layers heavy-tailed? Per-dimension kurtosis of the activation matrix $X$ predicts the kurtosis of random projections $r$ at $\rho = 0.97$ (panel~c). This is useful because per-dimension kurtosis is cheap to compute directly from activations, while signal kurtosis would require sampling many random projections; the tight correlation lets us diagnose formula breakdown from a single pass over $X$. Per-dimension kurtosis also predicts which layers fall below the curve in \Cref{fig:gamma_combined}: it correlates with formula overprediction at $\rho = -0.78$ for ESM3 and $\rho = -0.62$ for ESM2-650M (panels~d--e). Layers with near-zero kurtosis match the formula to within 1\%; layers with kurtosis above 0.5 show 5--9\% overprediction. The effect is concentrated in middle layers of ESM3 (layers 17--31, where $\kappa$ peaks at ${\sim}1.4$); early and late layers are near-Gaussian and match the formula closely (panel~f). ESM2-650M has elevated kurtosis only at its earliest layers ($\kappa > 3$ at layers 0--1), likely because the embedding layer's output is a token-vocabulary lookup, which produces a non-Gaussian distribution of activations across tokens.

The formula $\Phi(-C/\gamma)$ is therefore a useful upper bound on dead-by-ReLU rates and a close approximation whenever excess kurtosis is below ${\sim}0.3$, which covers most model-layer combinations in our dataset.

\section{Extended Analysis of Feature Revival Dynamics}
\label{app:recovery}

Section~\ref{sec:how-recovery-works} traces the recovery dynamics at
$\gamma = 20$ through two stages: a fast TopK revival driven by TopK
threshold collapse, followed by a slow dead-by-ReLU revival driven by
bias learning. The subsections below extend that picture across
$\gamma$, at the single-feature level, and through the bias-learning
bottleneck. Synthetic experiments use the SAE setup described in
\Aref{app:exp_setup}.

\subsection{Collateral death increases with $\gamma$}
\label{app:decomp_gamma}

\begin{figure}[H]
\centering
\includegraphics[width=\textwidth]{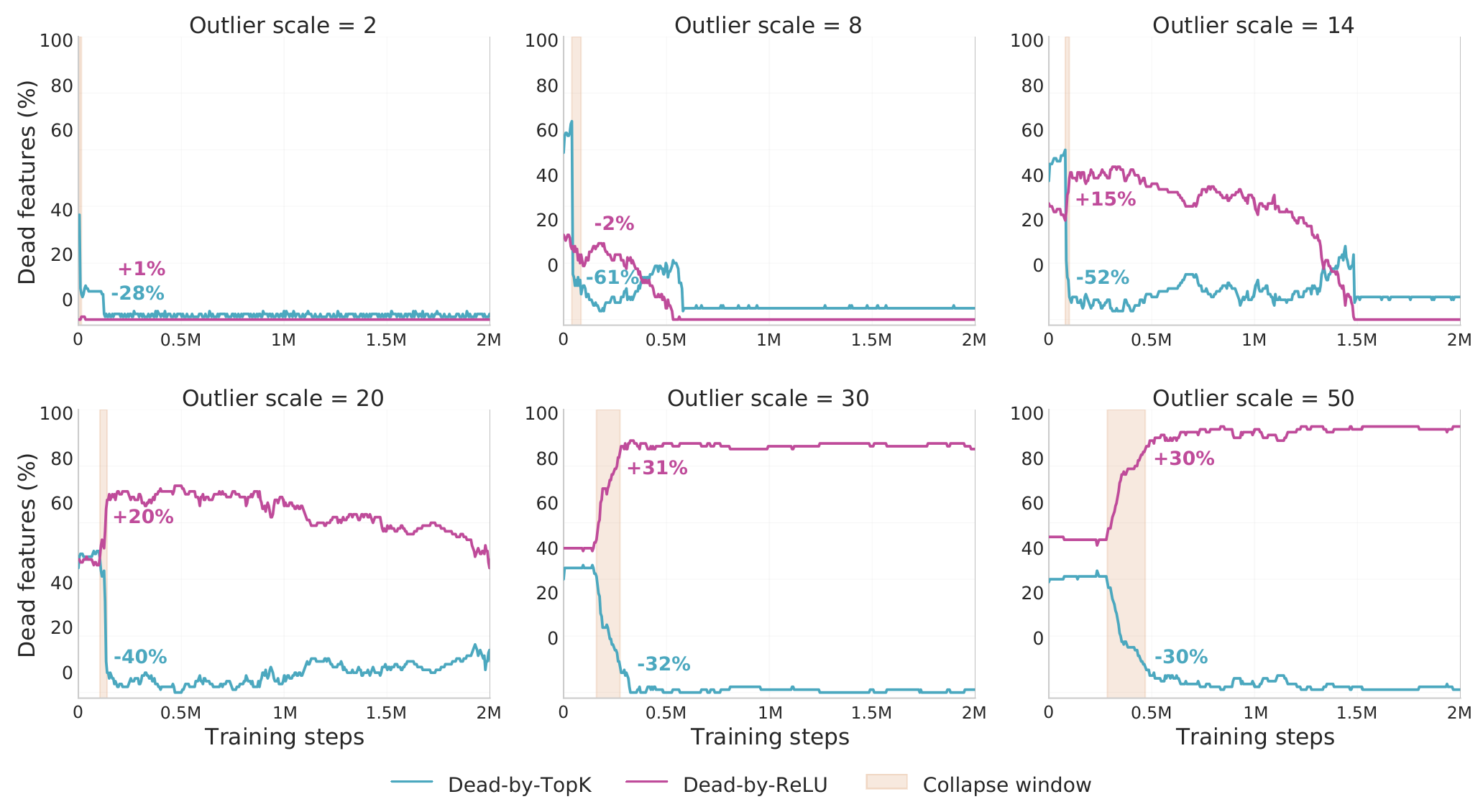}
\caption{\textbf{Collateral death increases with $\gamma$.} Dead-by-TopK
(cyan) and dead-by-ReLU (pink) over training at six $\gamma$ values. Orange
bands mark the collapse window (period of steepest TopK decline). Inset text
reports the change in each pathway during the collapse window and their ratio.
At low $\gamma$, TopK revival proceeds with minimal collateral death. At
$\gamma \geq 30$, the ratio approaches 1.0: new ReLU deaths roughly offset
TopK revival, explaining why total dead count in
Figure~\ref{fig:training}a appears flat at high $\gamma$ despite active
turnover.}
\label{fig:collateral_windows}
\end{figure}

To see how the two-stage structure varies with $\gamma$, we sweep
$\gamma \in \{2, 8, 14, 20, 30, 50\}$ and track dead-by-TopK and
dead-by-ReLU trajectories through the collapse window
(Figure~\ref{fig:collateral_windows}; collapse windows highlighted in
orange). Each panel reports the change in dead-by-TopK ($\Delta$TopK),
the change in dead-by-ReLU ($\Delta$ReLU), and their ratio during the
collapse window.

At $\gamma = 2$, the two-stage structure is barely visible. Dead-by-TopK
drops from ${\sim}$25 to near zero within the collapse window. Dead-by-ReLU
barely changes (ratio 0.04): almost no features cross below zero during
threshold collapse. At $\gamma = 8$, TopK revival is larger ($\Delta$TopK
$= -61\%$) but collateral death remains small (ratio $-0.03$): the threshold
drop opens slots without pushing many features below zero.

The picture changes at moderate $\gamma$. At $\gamma = 14$, dead-by-ReLU
rises by 15\% during the collapse window while dead-by-TopK drops by 52\%
(ratio 0.29). At $\gamma = 20$, the ratio reaches 0.50: half of the TopK
revival is offset by new ReLU deaths. This is why total dead count in
Figure~\ref{fig:training}a shows a smaller net drop at $\gamma = 20$ than
one might expect from the TopK revival alone.

At $\gamma \geq 30$, collateral death roughly equals TopK revival (ratio
${\sim}$1.0). Dead-by-ReLU rises by 31\% at $\gamma = 30$ while TopK drops
by 32\%. Total dead count appears flat not because nothing is happening, but
because TopK revival and collateral ReLU death are in approximate balance.

\begin{figure}[H]
    \centering
    \includegraphics[width=\textwidth]{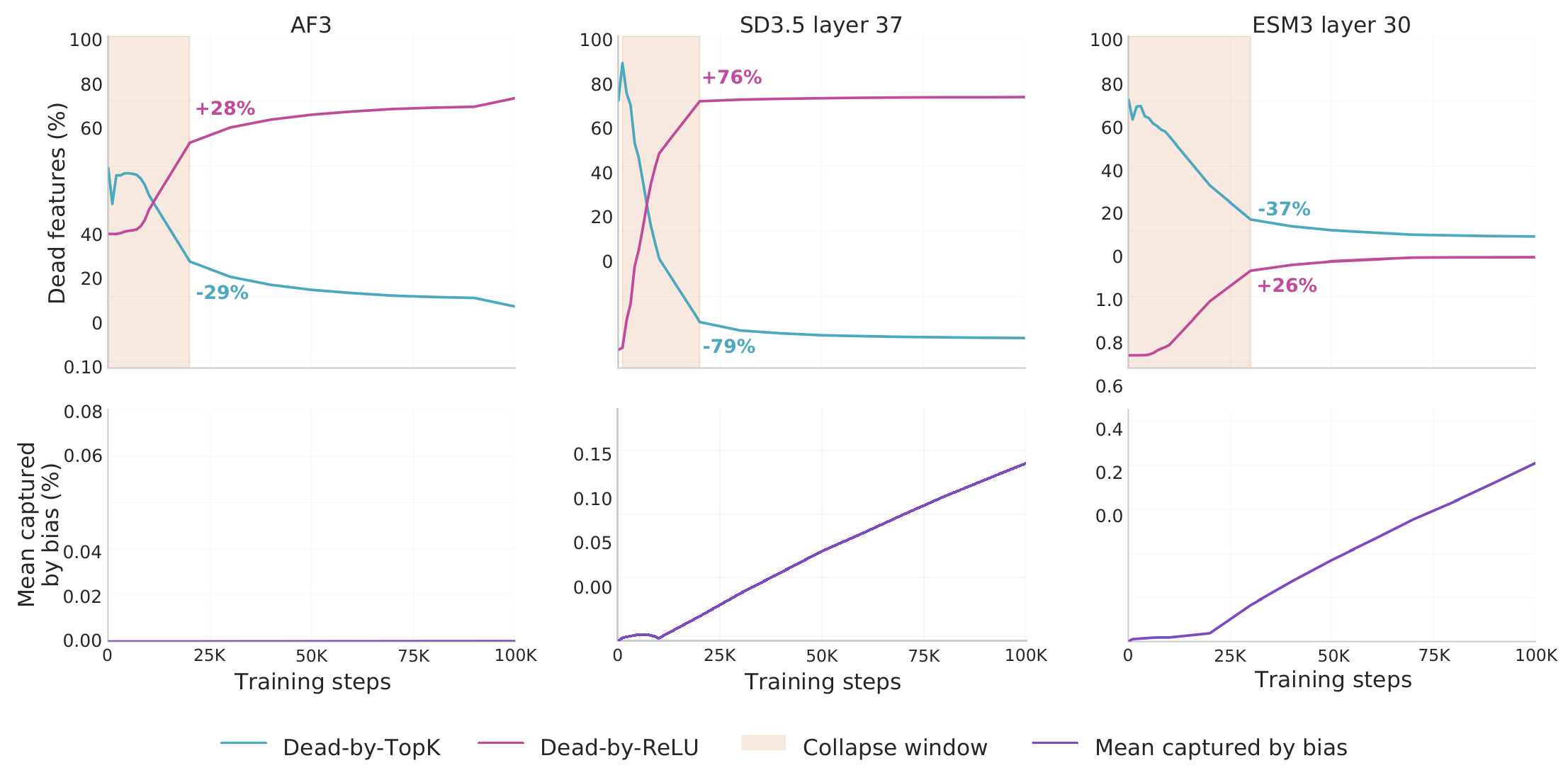}
    \caption{\textbf{The two-phase recovery dynamics from
    \cref{fig:training} hold on real models.} Dead-by-TopK (teal) and dead-by-ReLU (magenta) over training for three real models: AlphaFold3, Stable Diffusion 3.5 layer~37, and ESM3 layer~30. Shaded bands mark the collapse window. Top row: all three exhibit the same pattern seen in synthetic experiments: fast TopK revival during collapse (teal drops), simultaneous conversion of features to dead-by-ReLU (magenta rises), then a plateau. Annotated percentages show the change in each pathway during the collapse window. Bottom row: bias has captured less than 0.3\% of the activation mean after 100K steps in every case. These models have large mean offsets, so the required bias shift is correspondingly large, and bias learning is too slow to make meaningful progress within practical training budgets. Until the bias catches up, dead-by-ReLU features have no mechanism to revive, which is why they plateau. Dictionary size 8192, $k = 64$.}
    \label{fig:real_dynamics}
\end{figure}

The same dynamics appear on real models. \Cref{fig:real_dynamics} shows the
dead-by-TopK / dead-by-ReLU decomposition over training for three
high-$\gamma$ models: AlphaFold3, Stable Diffusion 3.5 layer~37, and ESM3
layer~30. All three exhibit the same pattern: fast TopK revival during
the collapse window, simultaneous conversion of features to dead-by-ReLU,
then a plateau. Collateral death is substantial in every case; on SD3.5,
dead-by-ReLU rises by 76 percentage points during collapse, nearly matching
the 79-point TopK drop. The bottom row shows the bottleneck directly: after
100K steps, the bias has captured less than 0.3\% of the activation mean in
every model. These real activations have much larger absolute scales than
our synthetic experiments at similar $\gamma$, so the required bias shift
is correspondingly large. Recovery time scales with the absolute
$\|\boldsymbol{\mu}\|$ (\Aref{app:lever_arm}), and bias learning is too
slow to make meaningful progress within practical training budgets. Until
the bias catches up, dead-by-ReLU features have no mechanism to revive,
which is why they plateau.  

\begin{figure}[H]
\centering
\includegraphics[width=0.8\textwidth]{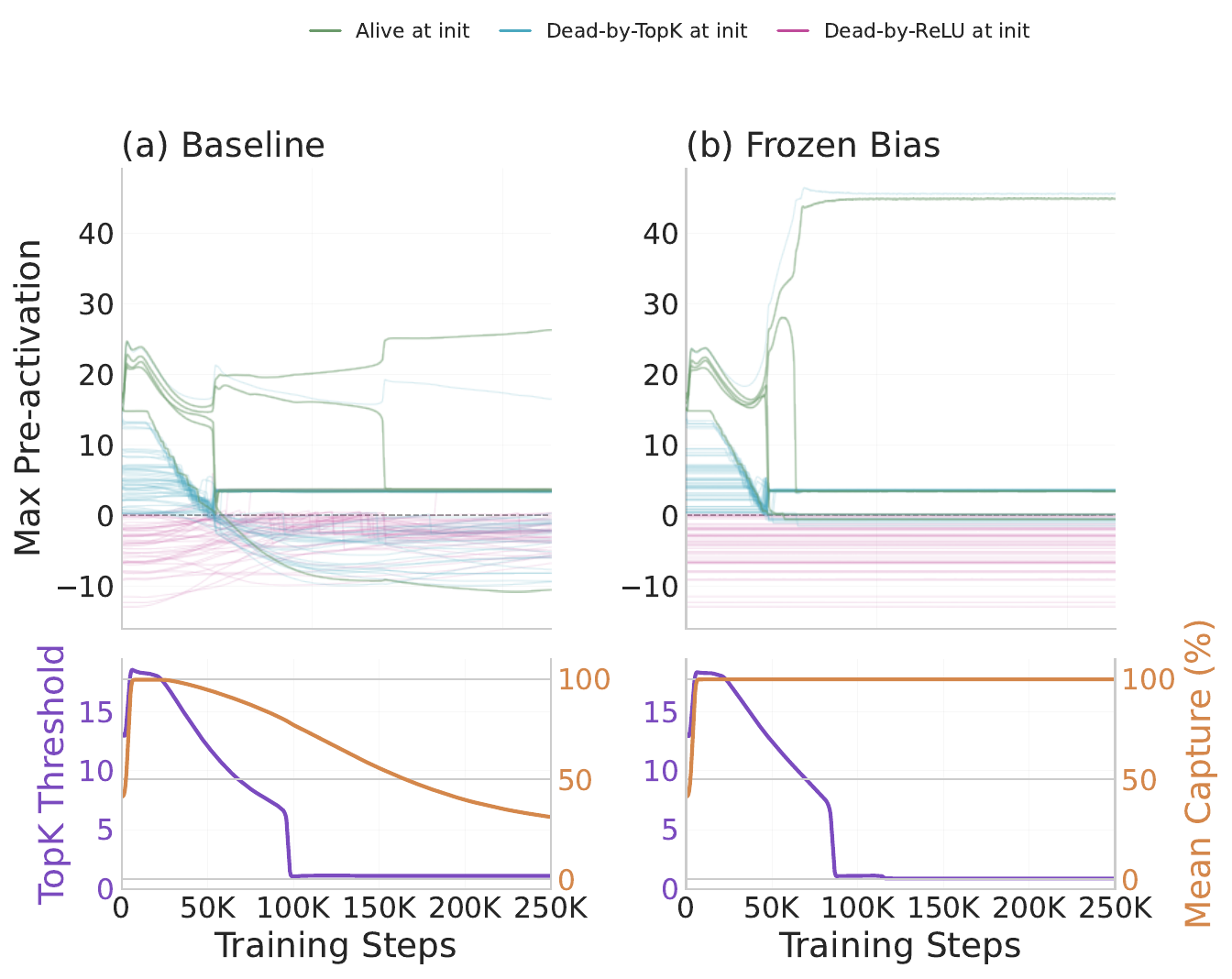}
\caption{\textbf{Per-feature pre-activation trajectories at $\gamma = 20$.}
Each line shows one feature's maximum pre-activation across evaluation inputs
over training. Colors indicate initialization status: alive (green),
dead-by-TopK (blue), dead-by-ReLU (pink). (a)~Baseline training. Dead-by-ReLU
features slowly rise toward zero as the bias learns $\boldsymbol{\mu}$.
(b)~Frozen bias. Dead-by-ReLU features are completely stationary: flat
horizontal lines throughout training. Bottom panels show the TopK threshold
(purple, left axis) and mean capture percentage (orange, right axis).
Threshold collapse coincides with features capturing the mean in both
conditions; dead-by-ReLU revival tracks bias learning in the baseline and
is absent when the bias is frozen.}
\label{fig:per_feature_traj}
\end{figure}

\paragraph{What drives collateral death.}
During threshold collapse, features strongly aligned with
$\boldsymbol{\mu}$ initially fire on every input and quickly learn to
reconstruct the mean. Their decoder columns are constrained to unit norm, so
they compensate by increasing their activation magnitudes. Once these features
capture $\boldsymbol{\mu}$, their activations decrease and the TopK
threshold drops, and previously excluded features can now enter the top-$k$.
But the same downward pressure that opens these slots also pushes some active
features' pre-activations below zero, converting them to dead-by-ReLU.
\Cref{fig:per_feature_traj} makes this visible on individual features.
The bottom panels show the TopK threshold (purple) rising as
mean-capturing features build up their activations, then collapsing
sharply once those features hit ${\sim}$100\% mean capture (orange) and
their activations start to decrease. The top panels (one line per
feature) show the consequence: alive features (green) follow the
threshold downward, and some borderline features cross below zero and
convert to dead-by-ReLU.

Higher $\gamma$ amplifies this effect. Mean-capturing features need to
reconstruct the full $\boldsymbol{\mu}$ using unit-norm decoder columns,
so they have to fire at activation magnitudes proportional to
$\|\boldsymbol{\mu}\|$. At higher $\gamma$, these activations start
larger, so when they later decrease (as the bias takes over) the drop is
correspondingly larger. The TopK threshold tracks these decreasing
activations and falls further, pushing more borderline features below
zero.

Bias learning can exacerbate collateral death. As the bias begins to learn
$\boldsymbol{\mu}$, it shifts all pre-activations, pushing features that
were already borderline (positive but decreasing) further below zero. This
effect is visible in Figure~\ref{fig:training}d and at the per-feature
level in \Cref{fig:per_feature_traj}: the dead-by-ReLU rise during the
collapse window is less pronounced when the bias is frozen (panel b) than
in the baseline (panel a). The effect is secondary to threshold collapse
itself, but it contributes at moderate $\gamma$ where many features hover
near zero.

\begin{figure}[H]
\centering
\includegraphics[width=0.6\textwidth]{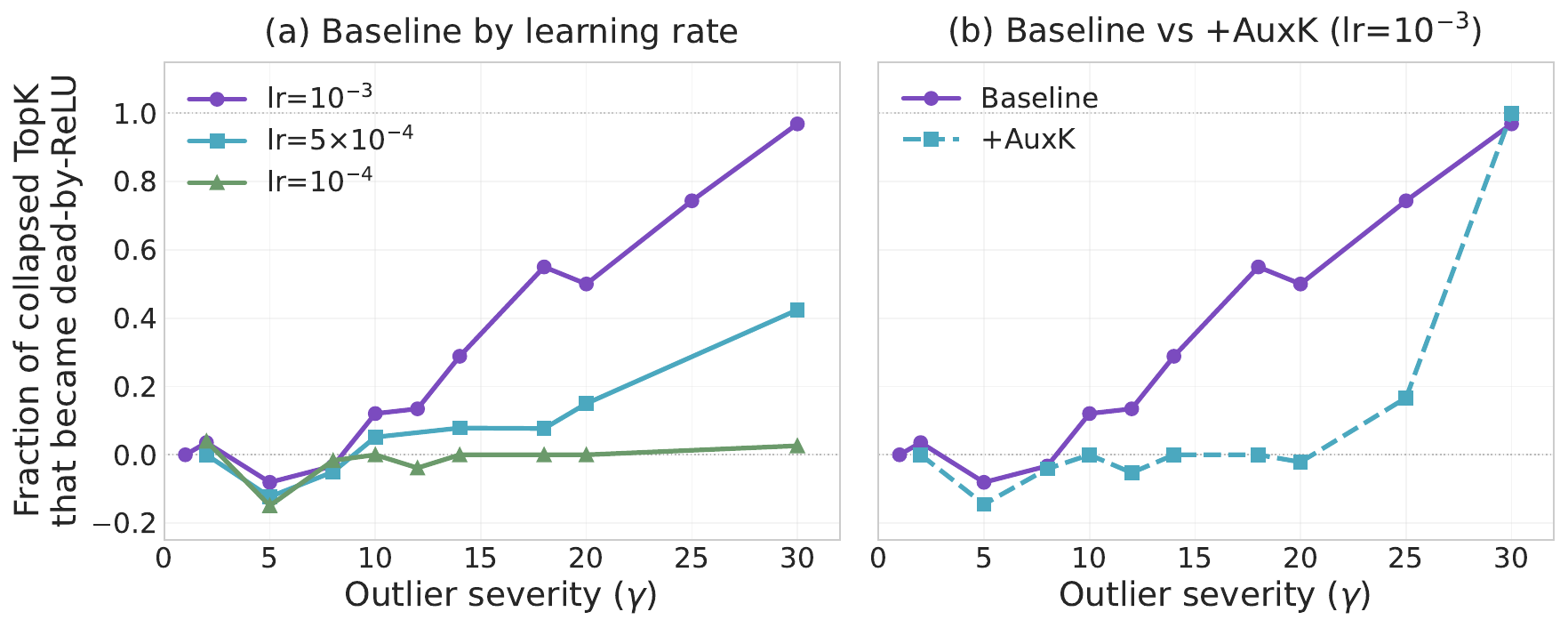}
\caption{\textbf{AuxK prevents collateral death at moderate $\gamma$ but
not at high $\gamma$.} The collateral death ratio (fraction of collapsed
dead-by-TopK features that become dead-by-ReLU during the collapse window)
as a function of $\gamma$. Baseline (purple): ratio rises steadily from
${\sim}$0 at $\gamma = 2$ to ${\sim}$1.0 at $\gamma = 30$. +AuxK (cyan
dashed): ratio stays near zero through $\gamma = 20$, then rises sharply
to ${\sim}$1.0 at $\gamma = 30$. Negative values at low $\gamma$ indicate
that some dead-by-ReLU features revived during the collapse window (net
decrease rather than increase).}
\label{fig:collateral_ratio}
\end{figure}

\paragraph{Collateral death and AuxK.}
Figure~\ref{fig:collateral_ratio} shows the collateral death ratio as a
function of $\gamma$, comparing baseline and +AuxK conditions. At moderate
$\gamma$ (5--20), AuxK keeps the ratio near zero: it provides gradient to
dead-by-TopK features during threshold collapse, helping them stabilize
above zero rather than crossing into dead-by-ReLU. At high $\gamma$
($\geq 25$), AuxK can no longer prevent collateral death: the threshold
drops are too steep and the pre-activation shifts too large. Both conditions
converge to a ratio near 1.0 at $\gamma = 30$.

This is the mechanism by which AuxK reduces dead-by-ReLU count in
Figure~\ref{fig:training}d, despite providing no gradient to features
with negative pre-activations. At moderate $\gamma$, preventing collateral
death is the difference between full recovery and permanent death. At high
$\gamma$, most dead-by-ReLU features were born dead at initialization, not
created as collateral, so this benefit addresses a shrinking fraction of
the total problem.

\paragraph{Learning rate sensitivity.}
The exact timing and magnitude of collateral death depend on learning rate.
Very low learning rates slow threshold collapse (features take longer to
capture the mean, so the threshold drops later and more gradually), reducing
collateral death. Very high learning rates produce steeper threshold drops
and more collateral death, because large weight updates can overshoot,
sending more features below zero in a single step. The qualitative trend
(collateral death increasing with $\gamma$) holds across the learning rates
we tested.

\subsection{Per-feature view of the two-stage recovery}
\label{app:per_feature}

\Cref{fig:per_feature_traj} shows per-feature pre-activation
trajectories at $\gamma = 20$ under baseline training (a) and frozen
bias (b). Each line is one feature's maximum pre-activation over
training, colored by its initial status (alive=green, dead-by-TopK=blue,
dead-by-ReLU=pink); bottom panels show the TopK threshold (purple) and
mean-capture progress (orange).

\paragraph{Without bias learning, dead-by-ReLU features stay dead.}
The pink features in baseline (a) slowly rise toward zero as the bias
learns $\boldsymbol{\mu}$. In frozen bias (b), the same features are
flat lines that do not move throughout training. Bias learning is the
only mechanism that revives them.

\paragraph{TopK collapse is bias-independent.}
In both panels, the TopK threshold collapses right after mean capture
completes (bottom panels), and the blue features jump upward during
this window on the same timescale (top panels). Frozen bias does not
prevent TopK revival.

\paragraph{The bias keeps pushing features down after they cross zero.}
In baseline (a), features that died during collapse don't stop there.
The bias keeps shifting them downward as it learns, so they end up well
into negative pre-activation territory where they're slow to revive. In
frozen bias (b), descending features stop near zero.

\subsection{Why bias learning is slow}
\label{app:lever_arm}

\begin{figure}[H]
\centering
\includegraphics[width=\textwidth]{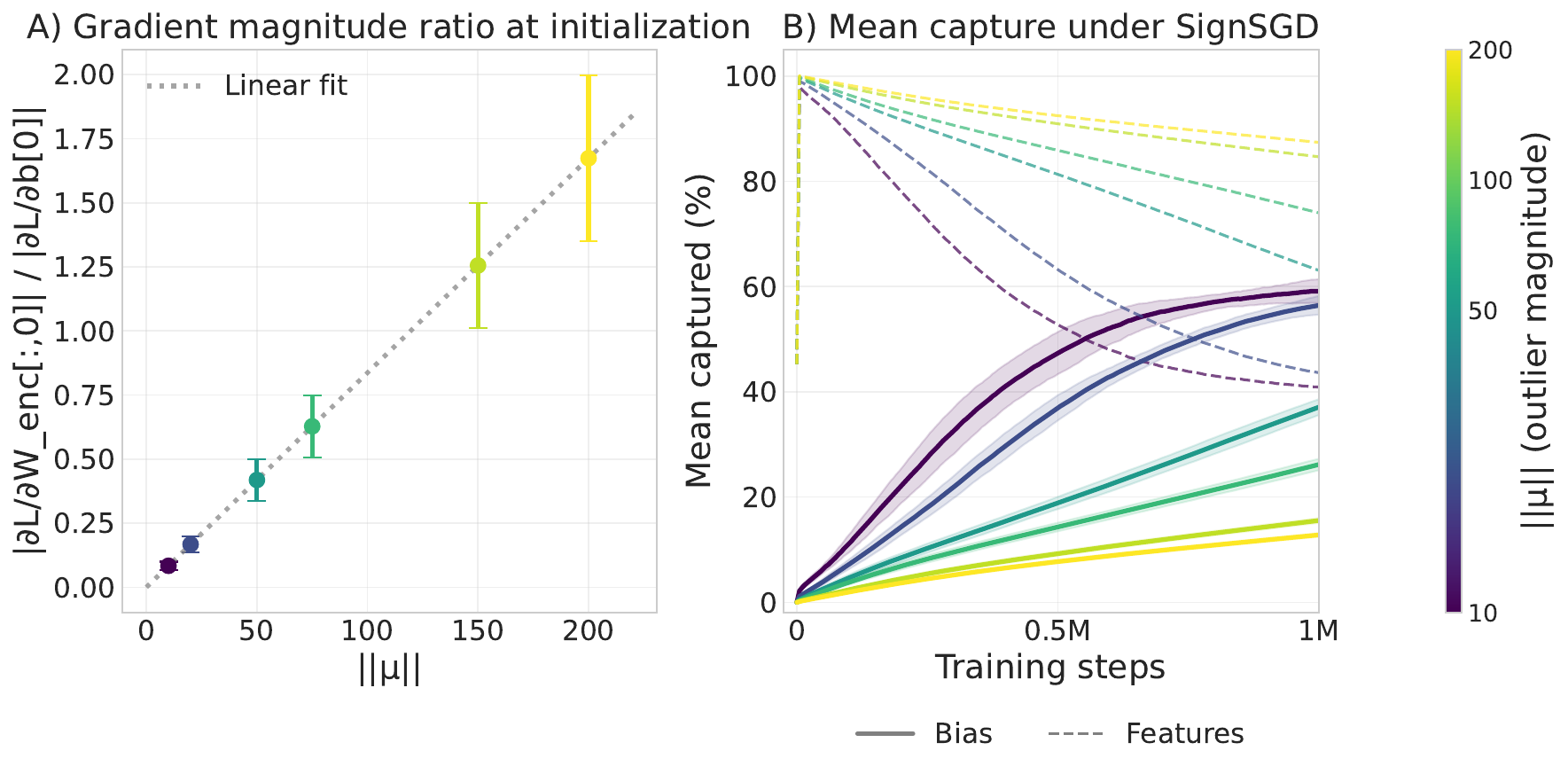}
\caption{\textbf{Feature gradients grow with $\|\boldsymbol{\mu}\|$;
bias gradients don't.} (a)~Gradient magnitude ratio (encoder weights on
outlier dimension vs.\ encoder bias) at initialization, plotted against
$\|\boldsymbol{\mu}\|$ at fixed $\gamma = 20$. The ratio scales
linearly with $\|\boldsymbol{\mu}\|$. (b)~Mean capture under SignSGD
(all parameters receive equal magnitude updates). Dashed lines: feature
capture (converges within a few thousand steps at all
$\|\boldsymbol{\mu}\|$). Solid lines: bias capture (depends strongly on
$\|\boldsymbol{\mu}\|$; at $\|\boldsymbol{\mu}\| = 10$ reaches
${\sim}$60\% by 1M steps; at $\|\boldsymbol{\mu}\| \geq 200$ barely
moves from zero). All conditions use $\gamma = 20$; colors indicate
$\|\boldsymbol{\mu}\|$ (viridis scale, 10 to 200).}
\label{fig:lever_arm}
\end{figure}

Two mechanisms make bias learning slow. The first is the asymmetric
gradient scaling noted in \Cref{sec:bias_bottleneck}: feature updates
pick up a factor of $\|\boldsymbol{\mu}\|$ from the input while bias
updates don't. We measure this directly at fixed $\gamma = 20$ with
$\|\boldsymbol{\mu}\|$ ranging from 10 to 200; the gradient ratio
$|\partial \mathcal{L}/\partial \mathbf{W}_\text{enc}[:,0]| \,/\,
|\partial \mathcal{L}/\partial \mathbf{b}_\text{enc}|$ at initialization
scales linearly with $\|\boldsymbol{\mu}\|$ (\Cref{fig:lever_arm}a). The
asymmetry persists under SignSGD \citep{bernstein2018signsgd}, where every parameter receives the
same magnitude update: features capture $\boldsymbol{\mu}$ within a few
thousand steps at all $\|\boldsymbol{\mu}\|$ values, while bias capture
depends strongly on $\|\boldsymbol{\mu}\|$ (at $\|\boldsymbol{\mu}\| =
10$ the bias reaches ${\sim}$60\% after 1M steps; at
$\|\boldsymbol{\mu}\| \geq 200$ it barely moves)
(\Cref{fig:lever_arm}b). The asymmetry is structural, not an optimizer
artifact.

\begin{figure}[H]
\centering
\includegraphics[width=0.6\textwidth]{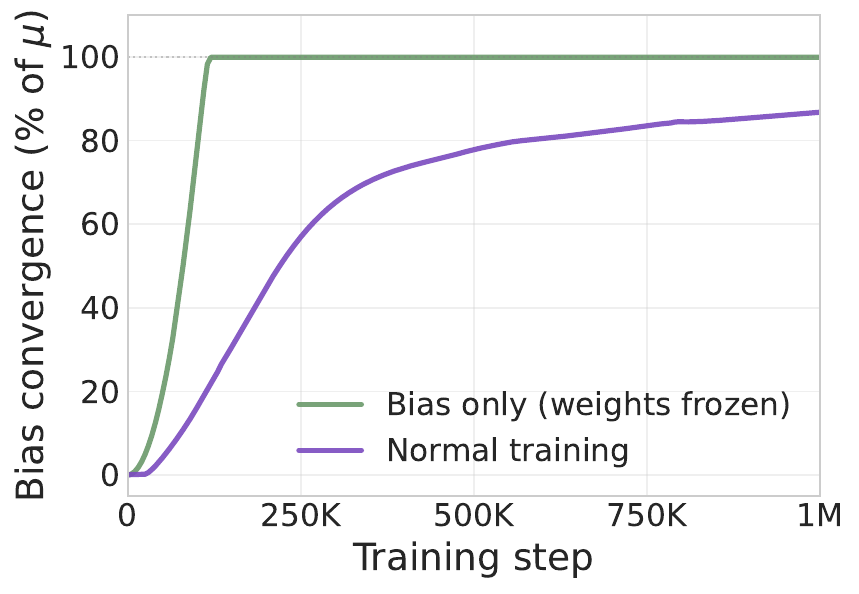}
\caption{\textbf{Feature competition slows bias learning.} Bias convergence
(fraction of $\boldsymbol{\mu}$ captured by $\mathbf{b}_\text{dec}$) at
$\gamma = 20$ under two conditions. Green: bias-only training (all weights
frozen): the bias converges to ${\sim}$99\% of $\boldsymbol{\mu}$ within
${\sim}$120K steps. Purple: normal training, where the bias reaches only
${\sim}$87\% after 1M steps, roughly 5$\times$ slower. When features
capture the mean first, they reduce the reconstruction residual that
drives bias learning.}
\label{fig:bias_competition}
\end{figure}

The second mechanism is gradient competition. Initial bias gradients
are large because every input has a large residual on the outlier
dimension, but features capture $\boldsymbol{\mu}$ within the first
${\sim}$100K steps. Once they do, the reconstruction residual shrinks
and the bias gradient loses its consistent direction. We can isolate
the effect by removing the competition: with all features frozen, the
bias is the only parameter that can reduce loss on the outlier
dimension, and it converges to ${\sim}$99\% of $\boldsymbol{\mu}$
within ${\sim}$120K steps. Under normal training (features learning in
parallel), it reaches only ${\sim}$87\% after 1M steps
(\Cref{fig:bias_competition}).

Together, these mechanisms decouple $\gamma$ from $\|\boldsymbol{\mu}\|$:
at fixed $\gamma$, the initial death rate is identical, but recovery
time scales with $\|\boldsymbol{\mu}\|$. AlphaFold3 layer~0 and ESM3
layer~9 both have post-LayerNorm $\gamma \approx 15$ and start at
${\sim}$98\% dead, but AlphaFold3's raw $\|\boldsymbol{\mu}\| \approx
80{,}000$ is 160$\times$ larger than ESM3's $\approx 500$. After the
same training budget, AlphaFold3 retains 98\% dead features while ESM3
drops to 83\%.

\begin{figure}[H]
\centering
\includegraphics[width=0.6\textwidth]{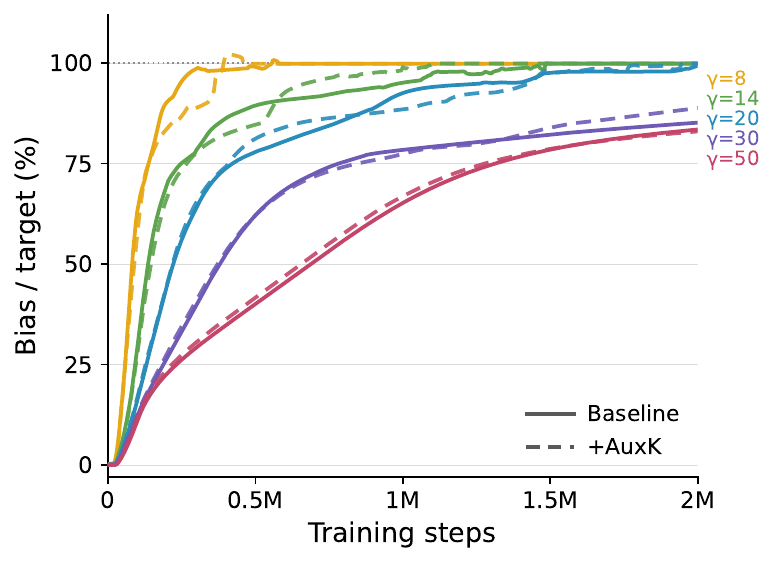}
\caption{\textbf{AuxK does not speed up bias learning.} Bias convergence
(fraction of $\boldsymbol{\mu}$ captured by $\mathbf{b}_\text{dec}$) over
training for baseline (solid) and +AuxK (dashed) at
$\gamma \in \{8, 14, 20, 30, 50\}$. At each $\gamma$ value, the two
conditions follow closely overlapping trajectories. The gap between
$\gamma$ values is far larger than the gap between baseline and +AuxK at
any given $\gamma$, confirming that outlier severity (not the presence of
auxiliary losses) determines the rate of bias learning.}
\label{fig:auxk_bias}
\end{figure}

\paragraph{AuxK does not speed up bias learning.}
Across $\gamma \in \{8, 14, 20, 30, 50\}$, the baseline and +AuxK bias
convergence trajectories overlap (\Cref{fig:auxk_bias}). The gap
between $\gamma$ values is far larger than the gap between baseline and
+AuxK at any fixed $\gamma$. AuxK's benefit on dead-by-ReLU operates
through encoder weight dynamics (stabilizing dead-by-TopK features
above zero during threshold collapse, \Aref{app:decomp_gamma}), not
through accelerating bias learning.

\section{Robustness across learning rates, architectures, and failure modes}
\label{sec:d_extended_res}

\subsection{Mean-centering makes synthetic models more robust across learning rates}
\begin{figure}[H]
\centering
\includegraphics[width=\linewidth]{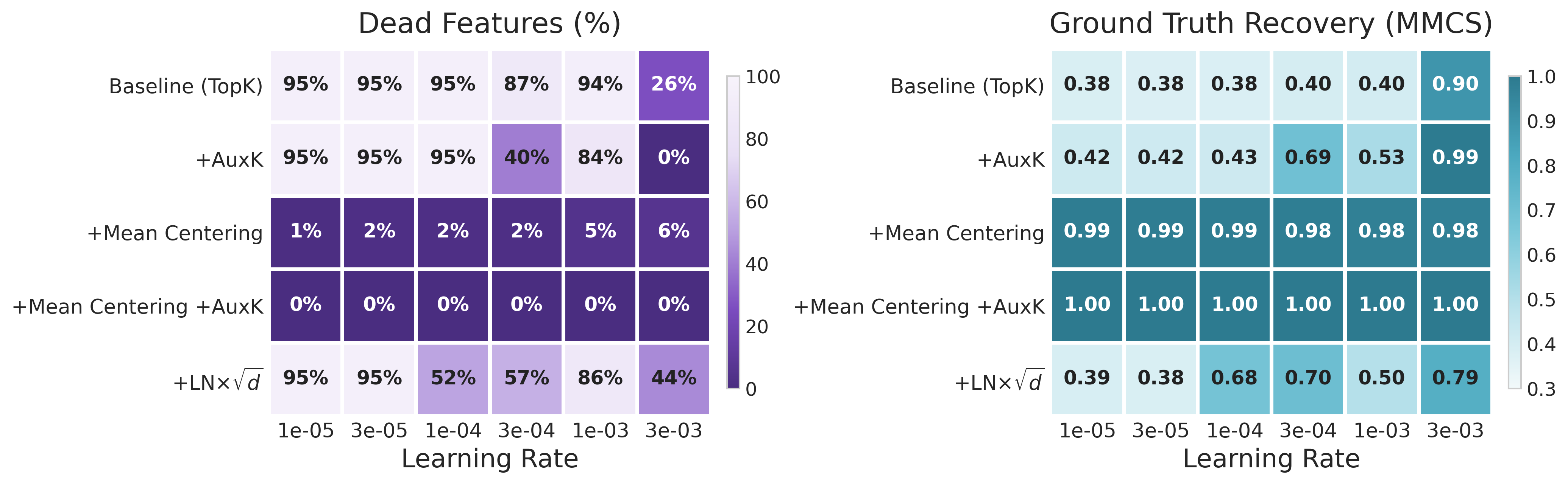}
\caption{\textbf{Mean-centering makes synthetic models robust to learning rate across two orders of magnitude.} Dead feature percentage (left) and ground-truth recovery via mean maximum cosine similarity (right) on synthetic data ($\gamma=40$), varying learning rate and preprocessing. TopK SAEs, dictionary size 8192, $k = 64$. Without mean-centering, the baseline and AuxK both show $>84\%$ dead features at all but the highest learning rate. Mean-centering brings death below $6\%$ and achieves near-perfect recovery ($\text{MMCS}\geq0.98$) across the full range. Layer normalization with $\sqrt{d}$ rescaling does not resolve death and hurts recovery.}
\label{fig:extended_learning_rates}
\end{figure}

\subsection{Mean-centering eliminates outlier-induced death across models and architectures}
\label{sec:app_mc_broad}

\paragraph{TopK SAEs.}

\begin{figure}[H]
\centering
\includegraphics[width=\linewidth]{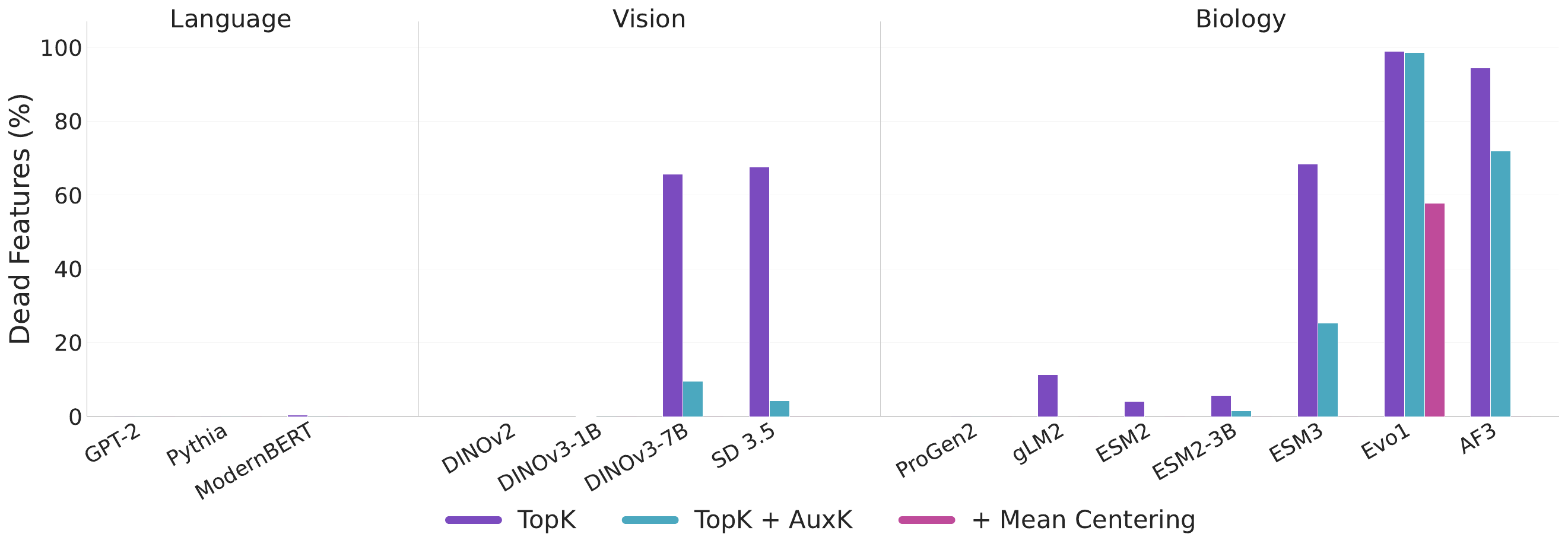}
\caption{\textbf{Mean-centering and AuxK each reduce death rates after training across the majority of embedding sources.} Single (middle-of-model) representative layer selected from each model. Dictionary size 8192, $k = 64$, 1M training steps. For each method--model pair we trained SAEs at several learning rates and report the run with the lowest reconstruction MSE.}
\label{fig:extended_all_models}
\end{figure}

Mean-centering eliminates outlier-induced death consistently across all models in our
evaluation suite (Figure~\ref{fig:extended_all_models}). Improvements are proportional to
outlier severity~$\gamma$: high-$\gamma$ models (ESM3, AlphaFold3, ESM2-3B) show the
largest reductions, while low-$\gamma$ models (GPT-2, Pythia) show minimal change. One
model, Evo1, is an exception whose residual death has a different geometric cause; we
analyze it in \Aref{sec:app_spectral}.

\paragraph{ReLU and JumpReLU SAEs.}

TopK SAEs enforce exact sparsity by keeping the $k$ largest latent
activations and zeroing the rest. ReLU and JumpReLU SAEs work differently:
they enforce sparsity indirectly, through their activation functions and
penalty terms. A ReLU SAE applies a standard ReLU to the encoder output
and adds an $L_1$ penalty on activations to discourage too many features
from firing at once. The competition between features is implicit: the
$L_1$ cost pressures the model to concentrate mass on fewer features
rather than explicitly capping how many can be active.
JumpReLU~\citep{rajamanoharan2024jumpingaheadimprovingreconstruction} replaces the smooth ReLU with a
hard threshold $\theta$: a latent is exactly zero if its pre-activation
falls below $\theta$ and passes through otherwise. Here sparsity is
governed by the threshold level (learned or scheduled) rather than a fixed
budget, so competition is again indirect: features must clear the
threshold to participate at all.

\begin{figure}[H]
    \centering
    \includegraphics[width=0.75\linewidth]{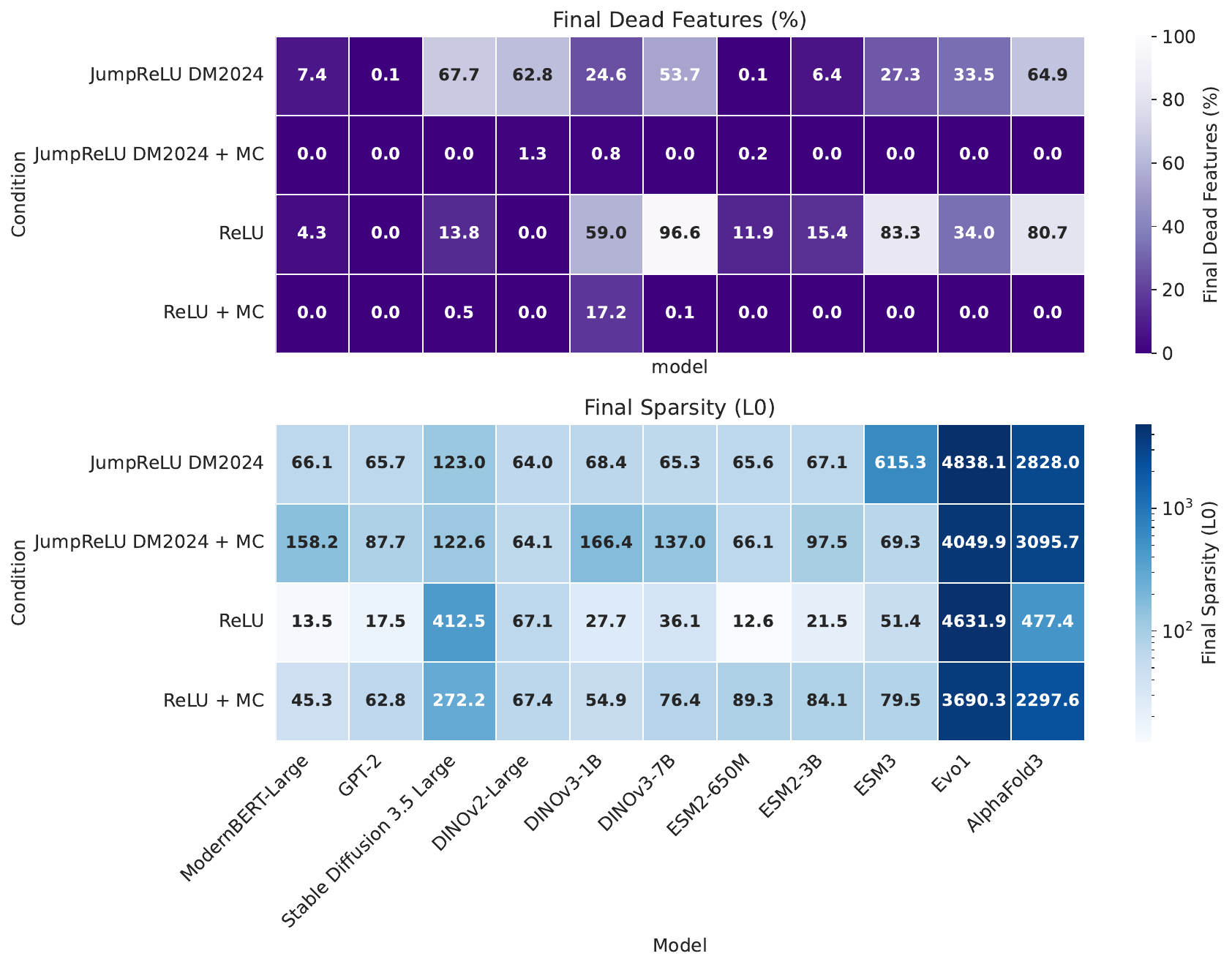}
    \caption{\textbf{Mean-centering reduces death across SAE architectures,
though sparsity control varies.} Top: dead features after 100K steps.
Bottom: final $L_0$. On extreme models (Evo1, AlphaFold3), JumpReLU+MC
achieves 0\% dead only by abandoning target sparsity ($L_0 > 640$). ReLU
sweep: $L_1 \in \{10^{-4} \text{--} 10^{-1}\}$, best $L_0$ near 64
selected. Dictionary size 8192.}
    \label{fig:relu-jumprelu-dead}
\end{figure}

A practical consequence is that the realized $L_0$ (average number of
nonzero latents) is harder to control than with TopK. Two runs with
different $L_1$ coefficients or threshold schedules can end up at very
different $L_0$ values, making apples-to-apples comparison difficult. We
target $L_0 \approx 64$ to match our TopK experiments, but for several
high-$\gamma$ models our hyperparameter sweep did not land in this range. We provide both the percent features that are dead features, as well as the final sparsity (L0) in Figure~\ref{fig:relu-jumprelu-dead}.

For JumpReLU, mean-centering helps substantially on models with moderate
$\gamma$: DINOv2-L drops from 62.8\% to 0.0\% dead at reliable $L_0$, and
SD~3.5 drops from 67.7\% to 0\%. On more extreme models (Evo1, AF3),
JumpReLU+MC achieves 0.0\% dead but only by abandoning sparsity entirely
($L_0 > 640$): the soft threshold cannot maintain target sparsity
on these inputs even with centering, and TopK's hard selection is needed.

\subsection{Mean-centering increases MonoSemanticity Scores}
\begin{figure}[H]
    \centering
    \includegraphics[width=\textwidth]{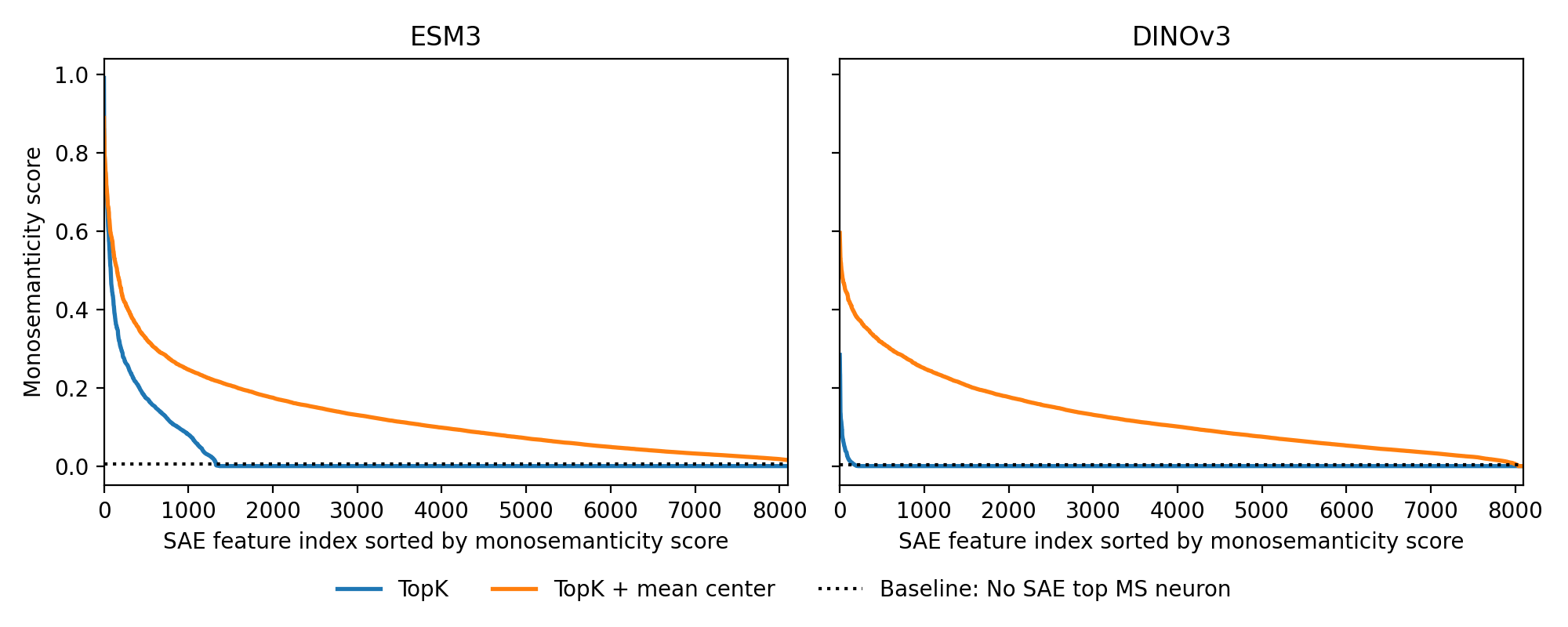}
    \caption{\textbf{Mean-centered SAEs produce more monosemantic features.}
    Monosemanticity Scores \citep{pach2025sparse} for all 8192 features, sorted by
    decreasing score. Each feature's MS measures the activation-weighted pairwise
    similarity of its top-activating inputs under an independent embedding model
    (ESM2 for ESM3; CLIP-ViT for DINOv3). Blue: baseline TopK. Orange: TopK + mean
    centering. Dotted: best single raw neuron (no SAE). On ESM3 (left), baseline
    features drop to zero around index 1500 (matching the ${\sim}1400$ alive features);
    mean-centered features maintain nonzero scores across roughly 5000 features. On
    DINOv3 (right), baseline features have almost no monosemanticity due to high death
    rates; mean-centering recovers meaningful scores across most of the dictionary.
    Dictionary size 8192, $k = 64$.}
    \label{fig:ms_score}
\end{figure}

\subsection{Mean-centering does not address training-dynamics sources of death}

The geometric analysis in the main text focuses on death that originates at initialization and persists into training. But features can also die from training dynamics alone, even when initialization is death-free. We describe three such mechanisms here.

\paragraph{High learning rate and high sparsity.}

\begin{figure}[H]
    \centering
    \includegraphics[width=0.5\linewidth]{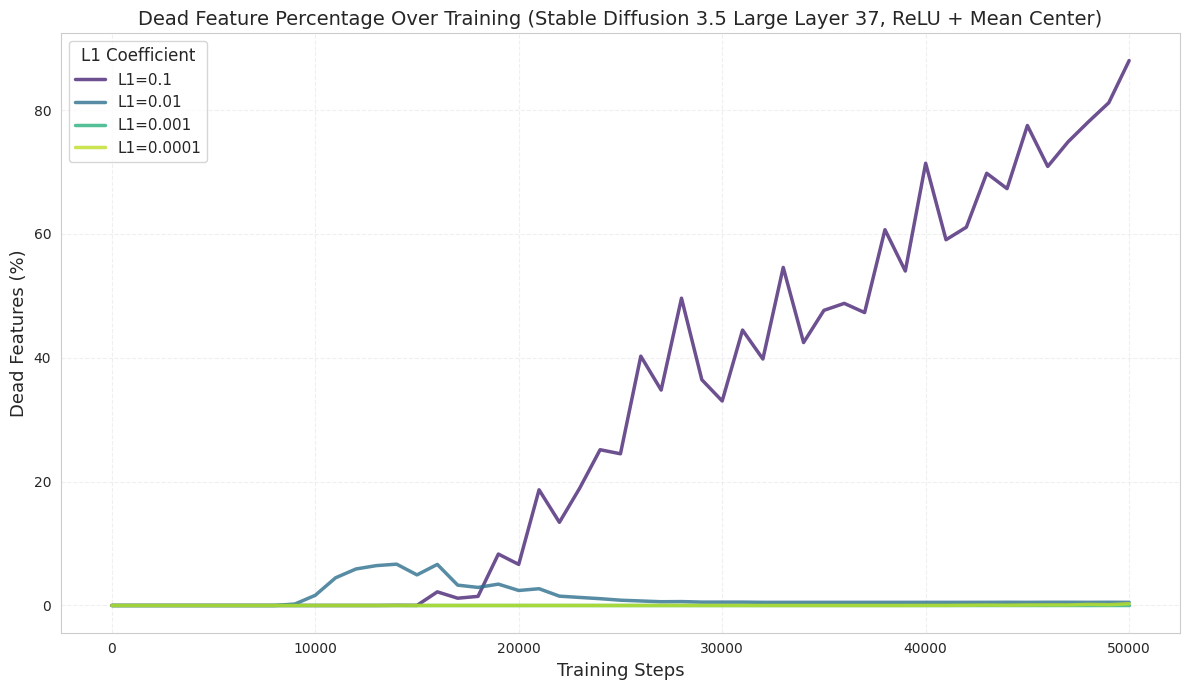}
    \caption{\textbf{High sparsity pressure causes death during training even
from a death-free initialization.} Dead features over training for a
mean-centered ReLU SAE (Stable Diffusion 3.5, layer~37) at four $L_1$
values. At $L_1 = 10^{-1}$, death rises to ${\sim}$80\% despite zero
geometric death at initialization.}
    \label{fig:dead_over_time_relu}
\end{figure}

\begin{figure}[H]
    \centering
    \includegraphics[width=0.75\linewidth]{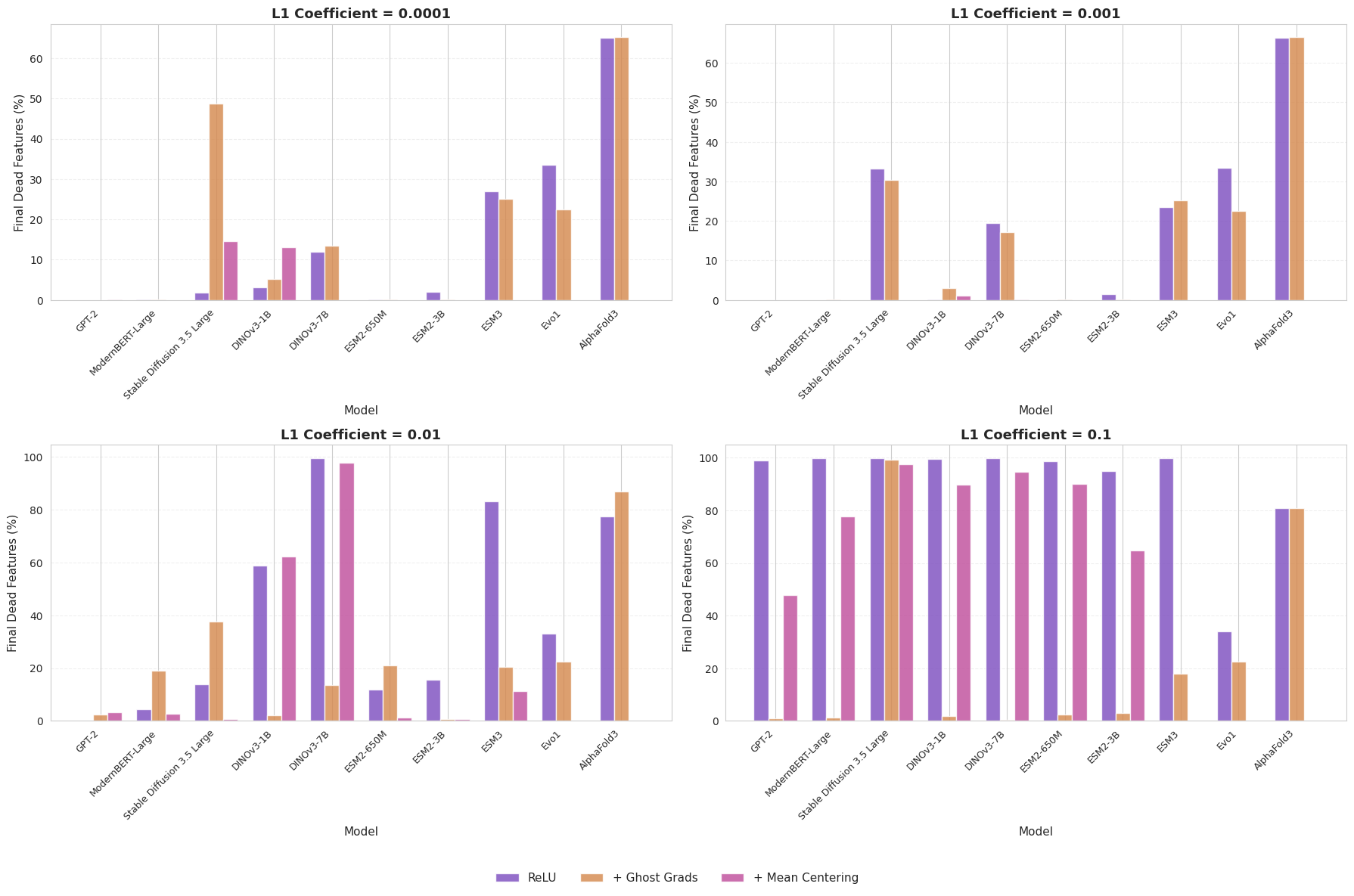}
     
\caption{\textbf{Ghost gradients and mean-centering address different failure
modes.} Dead features after 100K steps for ReLU SAEs (baseline, +ghost
gradients, +mean-centering) at $L_1 \in \{10^{-4} \text{--} 10^{-1}\}$.
Mean-centering fixes geometric death at low $L_1$; ghost gradients partially
help with training-dynamics death but not geometric death. Dictionary size
8192.}
    \label{fig:relu}
\end{figure}

Too-high learning rates can cause features to overshoot early in training and collapse to zero, regardless of initialization. Figure~\ref{fig:extended_learning_rates} shows this on synthetic data: learning rate alone drives death from near-zero to over 90\%, independent of whether geometric preprocessing was applied. High sparsity pressure has a similar effect. In ReLU SAEs, cranking up the $L_1$ coefficient progressively kills features that were initially alive, even when starting from a mean-centered, zero-death initialization (Figure~\ref{fig:dead_over_time_relu}). A decreasing-$k$ schedule in TopK SAEs can produce the same outcome.

Ghost gradients~\citep{jermyn2024ghost}, which inject synthetic gradient signal into inactive features, partially help on some models (Figure~\ref{fig:relu}). But they do not fix geometric death from mean offsets, much as AuxK does not on its own. We have not pinned down exactly why, but suspect a related mechanism: ghost gradients scale with the exponential of the pre-activation value, and when mean offsets push pre-activations far negative, the resulting gradient signal is tiny, likely too small to help the SAE learn to compensate for the offset.

\paragraph{Stale momentum.}

\citet{pmlr-v202-bricken23a} identified a separate optimizer-level failure mode: when a feature stays inactive for many steps, its Adam momentum terms go stale. If the feature then briefly activates, the accumulated stale momentum produces a disproportionately large update that can knock it right back to permanent inactivity. \citet{wang2025attentionlayersaddlowdimensional} showed that SparseAdam (a variant that only updates momentum for features with nonzero gradients) substantially reduces this kind of death during SAE training. We do not investigate SparseAdam in the present work, but note it as a complementary approach: mean-centering fixes the geometric conditions that cause death at initialization, while SparseAdam addresses the optimizer dynamics that can kill features over the course of training.

\subsection{Geometric median vs.\ arithmetic mean for centering}
\label{app:gm_vs_am}

We initialize the SAE pre-encoder bias to the geometric median (GM) of the activation distribution rather than the arithmetic mean (AM). On 15 of 19 models in our suite, GM and AM agree closely and the choice is immaterial. On the remaining four (DINOv3-B, DINOv3-7B, ModernBERT, ModernBERT-Large), a small fraction of outlier tokens with extreme activation norms inflates the AM by $1.7$--$5.5\times$ relative to the GM, and on three of these also rotates it sharply away from the typical token ($\cos(\bm{\mu}_\text{AM}, \bm{\mu}_\text{GM}) = 0.29$--$0.72$). Initializing the bias to this contaminated AM \emph{increases} init-time dead features on three of the four affected models, in some cases worse than no centering at all (\Cref{tab:gm-vs-am}). GM centering eliminates the problem ($\leq 2.3\%$ dead across all four).

The GM minimizes $\sum_i \|\mathbf{x}_i - \bm{\mu}\|_2$ rather than the sum of squared distances, giving it a breakdown point of $50\%$: up to half of token activations can be arbitrarily corrupted without affecting the estimate. Heavy-tailed token norms are common in vision transformers and recent encoder language models, where attention sinks and register tokens produce a small set of high-norm outliers. Trimmed means and the coordinate-wise median yield essentially identical results to GM on these models; we choose GM because it is affine-equivariant and requires no choice of trimming fraction. We compute it via Weiszfeld's algorithm on a calibration set of 500 tokens.

\begin{table}[h]
\centering
\small
\begin{tabular}{l c c c c}
\toprule
Model & Layer & \makecell{Dead \%\\(no center)} & \makecell{Dead \%\\(AM)} & \makecell{Dead \%\\(GM)} \\
\midrule
DINOv3-B & 3 & 57.6 & 91.7 & \textbf{0.0} \\
DINOv3-7B & 4 & 87.6 & 59.0 & \textbf{2.3} \\
ModernBERT & 16 & 55.1 & 89.0 & \textbf{0.0} \\
ModernBERT-Large & 20 & 34.5 & 88.3 & \textbf{0.0} \\
\bottomrule
\end{tabular}
\caption{\textbf{Geometric median centering eliminates outlier-induced death on models where arithmetic mean centering makes it worse.} TopK SAE, dictionary size 8192, $k = 64$, evaluated at random initialization on 100K held-out tokens. On three of four affected models, AM centering produces more dead features than no centering at all; GM centering brings dead-by-TopK below $2.5\%$ across all four. Layers shown are the layer with the largest gap between AM-centered and GM-centered death rates for each model.}
\label{tab:gm-vs-am}
\end{table}


\section{Features die when activation variance concentrates in too few directions}
\label{sec:app_spectral}

Mean-centering eliminates outlier-induced death across models (Section~6,
Figure~\ref{fig:extended_all_models}), but on some models and layers
substantial death remains at initialization even after centering. This
remaining death is entirely dead-by-TopK: centering removed the mean offset,
so no feature has permanently negative pre-activations, but many features
still never rank in the top~$k$ on any input.

At the representative mid-depth layers we train our models on
(\cref{fig:extended_all_models}), Evo1 is our only outlier where
substantial death survives mean-centering. Centering reduces init-time
death from 96\% to 73\% but does not eliminate it, and the residual does
not recover during training (${\sim}$58\% dead-by-TopK at the end). The fact that this remaining death is, like the outlier-induced death, present before any training implies that it is another quirk of the geometry of our activations.

When we look at post-centering death rates across every layer of every
model at initialization, several other models (DINOv3-7B, ESM2-3B, ESM3,
and ProGen2-base) also have nonzero post-centering death in a few transformer layers. These death rates are much smaller than Evo1's (20--25\%) so easier to recover to near zero during training and also primarily present in the earliest layers of the (typically worst at layer 1)
(\Cref{tab:spectral_fixes}), however they highlight that this phenomenon is not specific to Evo1 and provide more examples by which to study it.

We find that the common factor in each of these layers, most extreme in
Evo1, is that activation variance is highly concentrated in a small
number of principal components (PCs, the eigenvectors of the activation
covariance; their variances are the corresponding eigenvalues). When a
few PCs carry nearly all the variance, features that project onto them
dominate the TopK competition on every input, and the rest never fire.
We quantify this concentration with the \emph{effective rank} of the
covariance, $\exp(H(p_1, \ldots, p_d))$ where $p_i = \lambda_i / \sum_j
\lambda_j$ is the normalized eigenvalue distribution
\citep{roy2007effective}; this is low when variance is concentrated in
few PCs and high when it is spread across many. As with our other
metrics, we calculate this on post-LayerNorm activations to factor out
per-token magnitude differences, ensuring the concentration we measure
is directional rather than scale-driven.

\subsection{Variance concentration causes a small group of features to monopolize the TopK competition}
\label{sec:app_evo1_case}

This concentration of variance in a few PCs creates unequal competition among features, producing dead-by-TopK features. TopK selects the $k$ features with the
largest pre-activations on each input, so a feature only fires if its
pre-activation sometimes exceeds those of at least $n - k$ others. Each
feature's pre-activation is a dot product $\mathbf{w}_i^\top \mathbf{x}$,
so its variance across inputs is $\mathbf{w}_i^\top \Sigma \mathbf{w}_i$
for activation covariance $\Sigma$. This is the activation variance along
each PC of $\Sigma$, weighted by $\mathbf{w}_i$'s alignment with each PC.

When the per-PC variances are roughly equal, every random weight vector
captures similar total variance and features compete fairly. When the top
few PCs carry far more variance than the rest, only weight vectors
aligned with those top PCs capture meaningful variance; the rest produce
low-variance pre-activations and rarely rank in the top $k$. The few
features that happen to align monopolize the TopK competition on every
input, and the rest are dead-by-TopK.

We visualize this phenomenon using the activations of Evo1, where the effect is most severe.

\begin{figure}[H]
\centering
\includegraphics[width=0.9\textwidth]{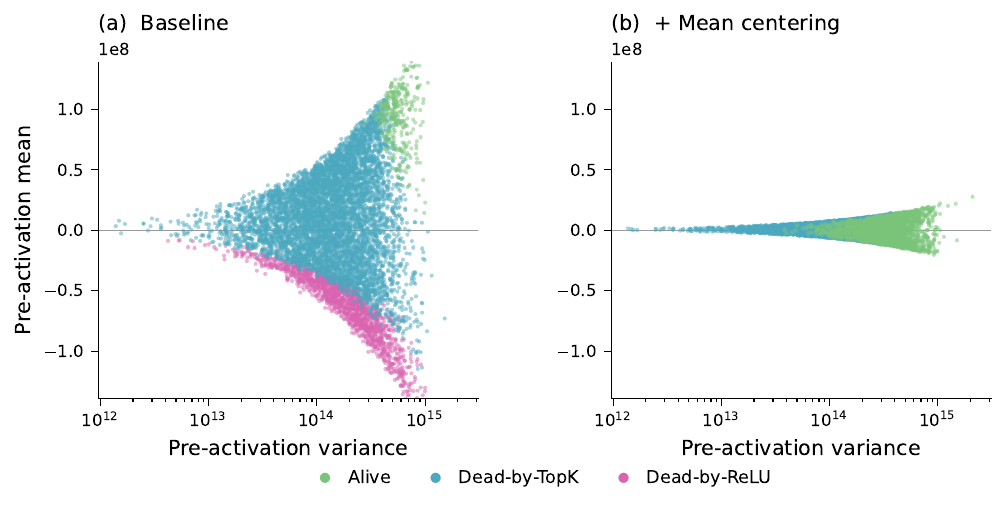}
\caption{\textbf{After mean-centering on Evo1, a minority of high-variance features monopolize the TopK competition.} \textbf{(a)}~Baseline (no preprocessing): pre-activation mean vs.\ pre-activation variance, colored by death type. 17\% dead-by-ReLU and 79\% dead-by-TopK.  \textbf{(b)}~Mean-centered: dead-by-ReLU disappears entirely. 73\% dead-by-TopK remains.   Dictionary size 8192, $k = 64$, evaluated at initialization on
  100K held-out samples.}
\label{fig:evo1_variance_ridgeline}
\end{figure}

In Figure~\ref{fig:evo1_variance_ridgeline}, we see that mean-centering removes the mean offset that produced dead-by-ReLU
features, yet 71\% of features remain dead-by-TopK, clustered in the
low-variance region of the scatter. The per-feature pre-activation
variances split sharply: alive features have $2.9\times$ higher median
variance than dead features.

However, variance alone does not fully predict which features die: the split is not a sharp cutoff. When most features'
pre-activations are driven by the same few underlying signals, a
lower-variance feature that tracks the same pattern as a dominant one,
just with smaller amplitude, peaks when the dominant one peaks and never
outranks it. Some moderate-variance features die for this reason; some
low-variance features survive by peaking on inputs where the dominant
features are middling.

\subsection{Removing variance concentration removes the death}
\label{sec:app_fixing_spectral}

If variance concentration causes the unequal competition, equalizing
variance should eliminate death. We evaluate two interventions: PCA
whitening (rotates and rescales the data) and Active Subspace
Initialization (changes only the SAE weights, leaving the data untouched).


\textbf{Equalizing the per-PC variances with PCA whitening eliminates the
death.} PCA whitening rotates activations into the PC basis and rescales
each component to unit variance. The resulting covariance is the
identity: every direction carries the same variance, so random weight
vectors capture the same expected pre-activation variance regardless of
orientation.

\begin{figure}[H]
\centering
\includegraphics[width=0.5\textwidth]{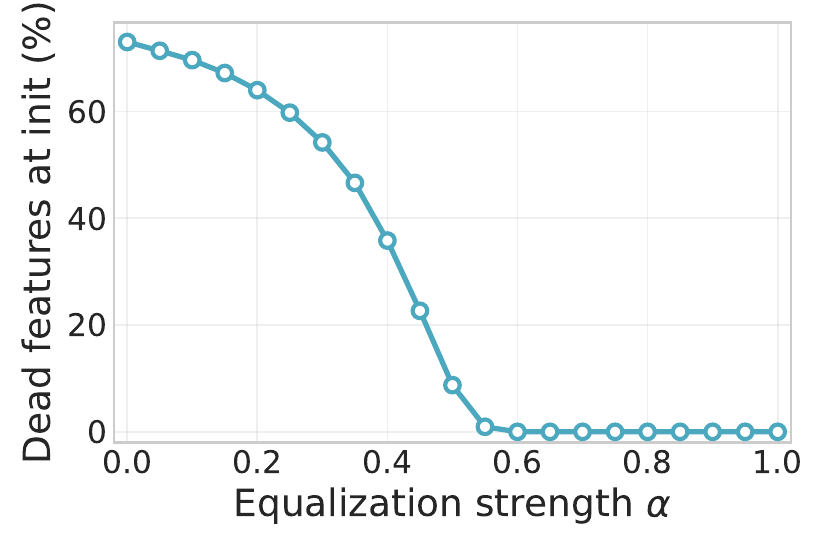}
\caption{\textbf{Smoothly equalizing the eigenvalue spectrum of Evo1
layer 14 activations monotonically reduces dead features.} Each
eigenvalue $\lambda_j$ is rescaled to $\lambda_j^{1-\alpha}$: $\alpha=0$
leaves the spectrum unchanged (mean-centering only), $\alpha=1$ sets
all eigenvalues to 1 (full PCA whitening). Death drops from 73\% to 0\%
by $\alpha \approx 0.6$. TopK SAE, dictionary size 8192, $k = 64$,
evaluated at random initialization on 100K held-out tokens.}
\label{fig:evo1_pc_whitening_curve}
\end{figure}

To understand how strongly eigenvalue concentration drives the death, we
smoothly interpolate between the original and equalized spectra.
Rescaling each eigenvalue $\lambda_j$ to $\lambda_j^{1-\alpha}$ for
$\alpha \in [0, 1]$ continuously deforms the spectrum from $\alpha=0$
(mean-centering only) to $\alpha=1$ (full PCA whitening). On Evo1
layer 14, dead-by-TopK features decrease monotonically with $\alpha$
and reach zero by $\alpha \approx 0.6$
(\Cref{fig:evo1_pc_whitening_curve}).

At $\alpha=1$, PCA whitening reaches 0\% dead features at initialization
on every model in our suite (\Cref{tab:spectral_fixes}). Z-scoring
(rescaling each coordinate to unit variance) does not fix the problem:
it touches only the diagonal of the covariance, leaving the off-diagonal
correlations that produce the eigenvalue concentration intact. On Evo1,
Z-scoring reduces death from 73\% to 49\% but does not reach zero.

Whitening is the simplest universal fix, but it changes the loss function:
dimensions that originally carried $1000\times$ more variance now contribute
equally to MSE, forcing the SAE to represent formerly low-variance
directions as faithfully as high-variance ones.
\citet{saraswatula_data_2025} found that this reweighting improves feature
quality on SAEBench. But high-variance directions may carry more of the
structure worth decomposing, so a less invasive fix is worth pursuing.


\textbf{Active Subspace Initialization fixes the remaining cases without changing the data.}
\citet{wang2025attentionlayersaddlowdimensional} observed a similar
phenomenon of variance concentration in the outputs of transformer attention blocks:
most of the variance lies along a small number of PCs resulting in low rank output matrices that have dead features when SAEs are trained on them, the same pattern
we see in our extreme cases. They propose Active Subspace Initialization
(ASI): project encoder and decoder weights onto the top $d_\text{init}$
principal components of the activations, where $d_\text{init}$ is a
hyperparameter that controls how many PCs are used (they find their
method is robust to its choice). This places features into the
high-variance subspace at initialization. Because the data is unchanged,
the loss function is identical to standard training.

The idea is that if features start in comparable-variance directions, the
TopK competition begins fair even though the underlying spectrum is
concentrated. This works in Wang et al.'s setting, where the attention
outputs in the models they study have effective ranks of 1000--2400 and many top PCs carry
comparable variance.

However, our activations are far lower-rank: in the extreme cases, one or two top PCs carry nearly all the variance. Any
$d_\text{init}$ in Wang et al.'s recommended range (their default is
768, far larger than our effective ranks of 3--31 on the affected
layers) still leaves a single dominant PC in the projected subspace,
and ASI does not meaningfully reduce dead features.

\input{rebuttal/gamma_472layers/spectral_fixes_table.tex}

The only $d_\text{init}$ values that produce low dead rates on Evo1 are 1 or 2, where ASI is degenerate: at $d_\text{init} = 1$ every
SAE feature is essentially $\pm \mathrm{PC}_1$, and at $d_\text{init} = 2$
all features live in a 2D plane regardless of dictionary size. A
high-capacity dictionary collapses to a 1D or 2D subspace at the start
of training, negating the benefit of using a large dictionary. 


\textbf{Practical guidance.} Most models reach near-zero dead features
under mean-centering alone (\Cref{tab:spectral_fixes}). When the post-LN
effective rank of the activation covariance is below ${\sim}2\%$ of the
hidden dimension, centering leaves substantial residual death and the
activations are intrinsically low-rank; PCA whitening reaches 0\% in
every such case. ASI is an alternative when changing the loss is
undesirable, with the caveat that small $d_\text{init}$ (on the order of
the effective rank of the affected layers) is needed since larger values
reintroduce the variance hierarchy. All diagnostics and fixes require
only a single pass over a batch of activations.

\end{document}

%% file: rebuttal/gamma_472layers/spectral_fixes_table.tex
\begin{table}[t]
\centering
\small
\begin{tabular}{l r r r r r l}
\toprule
Model & Layer & \makecell{Eff.\ rank\\$/\,d$ (\%)} & \makecell{Raw\\dead (\%)} & \makecell{MC\\dead (\%)} & \makecell{PCA\\dead (\%)} & Fix \\
\midrule
Evo1 & 14 & 0.07 & 95.9 & \textbf{73.0} & 0.0 & \textbf{PCA} \\
DINOv3-7B & 1 & 0.12 & 74.6 & \textbf{23.5} & 0.0 & \textbf{PCA} \\
ESM2-3B & 1 & 1.22 & 98.1 & \textbf{25.7} & 0.0 & \textbf{PCA} \\
ESM3 & 1 & 1.44 & 98.9 & \textbf{20.8} & 0.0 & \textbf{PCA} \\
ProGen2-base & 1 & 1.44 & 59.5 & \textbf{22.6} & 0.0 & \textbf{PCA} \\
\midrule
DINOv2-L & 1 & 1.30 & 55.8 & 2.3 & 0.0 & MC \\
SD3.5-L & 4 & 1.80 & 61.4 & 0.0 & 0.0 & MC \\
ProGen2-Large & 1 & 1.85 & 51.0 & 0.5 & 0.0 & MC \\
DINOv3-B & 1 & 2.85 & 86.4 & 1.0 & 0.0 & MC \\
DINOv2-B & 1 & 5.43 & 54.9 & 0.0 & 0.0 & MC \\
ESM2-650M & 1 & 7.92 & 91.6 & 0.0 & 0.0 & MC \\
AlphaFold3 & Pairformer & 8.13 & 99.0 & 1.0 & 0.0 & MC \\
ESM2-35M & 1 & 10.41 & 82.0 & 0.0 & 0.0 & MC \\
GLM2 & 27 & 19.70 & 49.0 & 0.0 & 0.0 & MC \\
ModernBERT-L & 25 & 19.75 & 51.3 & 0.0 & 0.0 & MC \\
Pythia-410M & 18 & 31.09 & 10.3 & 0.0 & 0.0 & MC \\
Pythia-70M & 2 & 37.49 & 3.6 & 0.0 & 0.0 & MC \\
ModernBERT & 5 & 37.63 & 67.5 & 0.0 & 0.0 & MC \\
GPT-2 & 11 & 45.53 & 32.7 & 0.0 & 0.0 & MC \\
\bottomrule
\end{tabular}
\caption{\textbf{Per-model dead-feature rates at the worst (highest MC dead) layer.} Raw, MC, and PCA columns are dead-by-TopK percentages at initialization. Effective rank is computed post-LayerNorm and normalized by hidden dimension $d$. The \textbf{Fix} column gives the recommended preprocessing per model: MC when post-MC death is below 5\%, PCA otherwise.}
\label{tab:spectral_fixes}
\end{table}